\documentclass[twoside]{article}

\usepackage{PRIMEarxiv}

\usepackage[utf8]{inputenc}     
\usepackage[T1]{fontenc}        
\usepackage{multicol}

\usepackage{microtype}          
\usepackage{setspace}           
\usepackage{textcomp}           
\usepackage{color,soul}         
\usepackage{lipsum}             

\usepackage{amsmath, amssymb, amsfonts, mathtools} 
\usepackage{eucal}              
\usepackage[version=4]{mhchem}             
\usepackage{nicefrac}           
\usepackage{nccmath}

\usepackage{graphicx}           
\graphicspath{{media/}}         
\usepackage{caption, subcaption}
\usepackage{float}              
\usepackage{adjustbox}          
\usepackage{booktabs}           
\usepackage{makecell}
\usepackage{multirow}           
\usepackage{longtable}          
\usepackage{rotating}           
\usepackage{pdflscape}          
\usepackage{placeins}           

\usepackage{algorithm}          
\usepackage{algpseudocode}      

\usepackage{enumitem}           

\usepackage{hyperref}           
\usepackage{url}                

\usepackage{fancyhdr}           

\usepackage{natbib}             

\usepackage{xpatch}             

\title{Discovering Interpretable Ordinary Differential Equations from Noisy Data
}

\author{
  Rahul Golder\footnotemark[1] \\
  \texttt{rahulgolder8420@tamu.edu}
  \And
  M. M. Faruque Hasan \\
  \texttt{hasan@tamu.edu}\footnotemark[1]\thanks{Texas A\&M Energy Institute, Texas A\&M University, College Station, TX 77843, USA} \thanks{Corresponding author: \texttt{hasan@tamu.edu}} \\
}

\begin{document}
\maketitle

\begin{abstract}
The data-driven discovery of interpretable models approximating the underlying dynamics of a physical system has gained attraction in the past decade. Current approaches employ pre-specified functional forms or basis functions and often result in models that lack physical meaning and interpretability, let alone represent the true physics of the system. We propose an unsupervised parameter estimation methodology that first finds an approximate general solution, followed by a spline transformation to linearly estimate the coefficients of the governing ordinary differential equation (ODE). The approximate general solution is postulated using the same functional form as the analytical solution of a general homogeneous, linear, constant-coefficient ODE. An added advantage is its ability to produce a high-fidelity, smooth functional form even in the presence of noisy data. The spline approximation obtains gradient information from the functional form which are linearly independent and creates the basis of the gradient matrix. This gradient matrix is used in a linear system to find the coefficients of the ODEs. From the case studies, we observed that our modeling approach discovers ODEs with high accuracy and also promotes sparsity in the solution without using any regularization techniques. The methodology is also robust to noisy data and thus allows the integration of data-driven techniques into real experimental setting for  data-driven learning of physical phenomena.
\end{abstract}

\keywords{Model Identification \and Machine Learning \and Inverse modeling}

\vspace{1 cm}

\section{Introduction}

The governing principles of many systems can be often described using ordinary differential equations (ODE). Given a set of input-output data ({$\mathbf{\hat{x}}$} vs. {$\mathbf{\hat{y}}$}) from a system, we would like to find the governing equation that dictates the system behavior. We assume this governing equation to be a homogeneous linear ODE of $P^{th}$ order with coefficients $C_{p}(p = 0, \cdots, P)$ as follows:

\begin{ceqn}
    \begin{equation}
\label{sec:methodology:eq:general_ODE}
    \sum_{p=0}^{P}C_{p}\frac{d^{p}y}{dx^{p}} = 0.
\end{equation}
\end{ceqn}

Our goal is to identify the unknown coefficients of Eq. \ref{sec:methodology:eq:general_ODE} that best represents the data. Discovering governing equations from experimental or simulated data describing the underlying phenomena is an active area of research. There are many systems of practical importance where the actual physics is only partially known (grey-box) or completely unknown (black-box). Assumptions are often made to develop approximate mathematical models to describe and predict the system behavior.
Two approaches are predominant for model building, namely function-based and data-driven approaches. Function-based approaches signify determination of a functional form that fits the data and closely follows the underlying governing equation. The selection of the functional form is often driven by fundamental understanding of the system dynamics and knowledge of physicochemical phenomena (e.g. reaction, phase separation, adsorption, etc.), mass and energy conservation laws, thermodynamics (e.g. chemical potential, fugacity), and transport.
 Data-driven approaches, such as machine learning (ML) and Artificial Neural Networks (ANNs), on the other hand, do not require such phenomenological understanding and consider the system as a black box.
 

Lahouel et al. \cite{lahouel2024learning} introduced a non-parametric method for discovery of ODEs describing time variant systems using Reproducing Kernel Hilbert Spaces (RKHS). Although RKHS ensures existence and uniqueness of ODE solutions, its performance can be sensitive to the choice of appropriate kernels used in RKHS. Venkatramana \cite{venkataraman2011determining} posed equation discovery as a ``regularized inverse problem" and approximated the true model using a set of polynomial basis functions. This approach depends on the quality of numerical integration, and the performance degrades in the presence of noisy data. Wahba et al. \cite{wahba1977practical} addressed the problem of solving ill-posed linear operator using Tikhonov regularization in a RHKS space. Using this approach, linear inverse problems are solved, which correlates to solving a least square regression problem for estimating the coefficients of a generalized ODE. Long et al. \cite{long2024equation} introduced a Bayesian framework that quantifies uncertainty for sparse selection of operators and estimates target functions and their derivatives in a noise-robust manner. 
Recently, Sun et al. \cite{sun2022bayesian} proposed Bayesian spline learning (BSL) that uses alternating direction optimization (ADO) to train a spline model on sparse data and performs derivative computation. Additionally, using stochastic weight averaging gaussian (SWAG), posteriors are approximated over the governing equation coefficients. 

Symbolic regression has emerged as an alternative tool for model discovery. Schmidt et al. \cite{schmidt2009distilling} introduced ``Eureqa" for discovering symbolic equations directly from experimental data. It uses a genetic programming \cite{mirjalili2019genetic} based approach for identifying free-form mathematical expressions without requiring a predefined library, emphasizing interpretable discovery rather than black-box predictions. However, scalability becomes an issue for high dimensional and multi-physics systems. 
ALAMO \cite{wilson2017alamo} finds approximate general solution of the data by optimally combining various functional building blocks using mixed-integer nonlinear optimization (MINLP).
Brunton et al. \cite{brunton2016discovering}, in their seminal work of SINDy (Sparse Identification of Non-linear Dynamics) performed sparse prediction of a time-varying governing equation using sequentially thresholded least squares (STLSQ) algorithm. 
SINDy and its variants require a feature library which contains multiple basis functions. Through learning the sparse coefficient set of these basis functions, a mathematical description of a physical system is discovered.
In spite of its success, SINDy suffers from the following issues. It uses a high dimensional feature library for sparse prediction which increases the computational complexity of the model. Additionally, it computes the derivatives directly from the data, which may be sensitive to noise.
Wentz et al. \cite{wentz2023derivative} proposed derivative based SINDy (DSINDy) to tackle the issue of noisy data. 
Lejarza et al. \cite{lejarza2022data} posed the equation discovery as a moving horizon optimization (MHO) problem. This framework solves a nonlinear program (NLP) over moving time window followed by basis function construction and structured optimization to identify active terms that contribute to the system dynamics. 
Shi et al. \cite{shi2025compressive} integrated compressive sensing and mixed integer optimization to improve the robustness and accuracy of symbolic regression. They employed a two-stage process where a convex sparse regression problem is solved to identify a subset of basis functions followed by mixed-integer quadratic (MIQP) optimization to determine optimal combination of basis functions and their corresponding coefficients.

Recently, ANNs have shown great promise for model approximation. 
AI-Feynman \cite{udrescu2020ai} utilizes the approximation capability of feed-forward networks. Following the approximation, a recursive polynomial fit is used to determine a closed form of the function. 
The same group proposed a modified version of the algorithm as AI-Feynman v2 incorporating a pareto-optimal frontier between accuracy and complexity to improve the noise robustness and scalability of the algorithm \cite{udrescu2020ai2}. However, the symbolic brute-force search used in this framework is limited by fixed set of alphabet operators. Additionally, dimensional analysis used to reduce the number of variables assumes accurate unit vectors which may not be available in a real world dataset. 
Greydanus et al. \cite{greydanus2019hamiltonian} proposed Hamiltonian neural networks (HNNs) for learning physical systems by leveraging hamiltonian mechanics. In this approach, Hamiltonian parameterization is learned as an ANN which maps position and momentum to scalar energy. HNNs learn from the observed data without any supervision of the conserved quantities. 
Cranmer et al. \cite{cranmer2020lagrangian} proposed Lagrangian neural networks (LNNs) that operates on arbitrary coordinates to encode the system dynamics using Euler-Lagrange formulation parameterized by an ANN. LNNs improve over HNNs in conserved energy systems over long simulation period. However, training can be sensitive to ill-conditioned Hessians for which pseudoinverses and regularizations are required.

While many of the techniques mentioned above are able to find surrogate     models that fit the data well, these models often lack physical meaning and interpretability. Another limitation is that they lack extrapolation beyond the training domain. These approaches are also not suitable in the presence of noisy data. Noise and measurement error are common in real systems. However, proposing an appropriate functional form for noisy data is a challenge. 

To address these gaps, we develop an approach for discovering physically meaningful governing equations such as ODEs that have a realistic form and general solution that resembles real-world applications. It consists of three major steps: (i) identifying the parameters of a general solution of the governing ODE from noisy data, (ii) polynomial approximation of the general solution to obtain derivative information, and (iii) solving a linear system for coefficient estimation of the ODE. 
To address the issue of noisy data, we first estimate the parameters of a prepostulated general solution of the ODE in Eq. \ref{sec:methodology:eq:general_ODE}.
For efficient polynomial approximation of the smooth functional representation, we use a B-spline model with adaptive knot selection. A gradient matrix is calculated from the analytical spline. Because of the polynomial nature of the spline, we are able to obtain linearly independent derivatives.
Finally, a linear coefficient estimation is used to compute the null space of the gradient matrix which indicates the coefficients of the underlying governing equation. 
The approach is demonstrated using two test cases, namely the spring-mass system, and the photolytic degradation of estrogen disrupting chemicals (EDCs). The spring mass system represents a general class of harmonic oscillator systems, where the response of the system changes depending on the system property and the experimental setup.

The paper is organized as follows: Section \ref{sec:methodology} provides detailed description of the proposed methodology. Section \ref{sec:results} provides the results from the computational experiments using two case studies. Finally, Section \ref{sec:conclusion} includes our concluding remarks.

\section{Methodology}
\label{sec:methodology}
\begin{figure}[htbp]
    \centering
    \includegraphics[width=\linewidth]{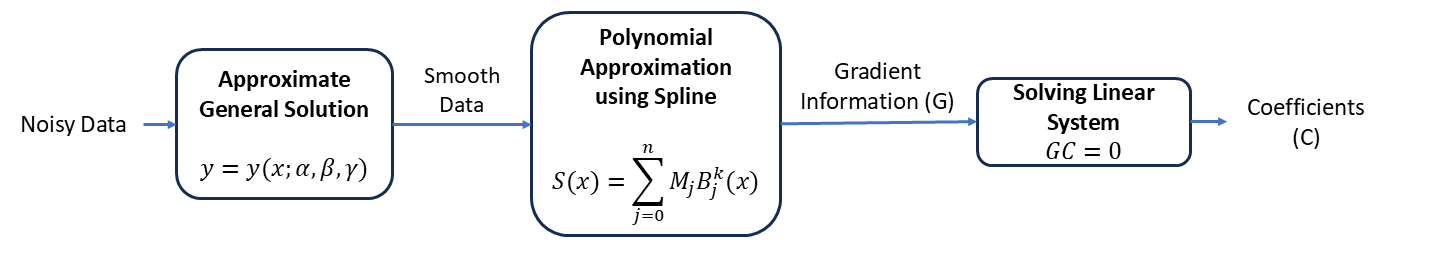}
    \caption{Proposed Methodology.}
    \label{fig:flowchart}
\end{figure}

The overall ODE discovery workflow is shown in Figure \ref{fig:flowchart}.
Approximate general solution step computes a high fidelity functional form of the present noisy dataset to create a smooth representation. The approximate general solution is postulated using the same functional form as the
analytical solution of a general homogeneous, linear, constant-coefficient ODE. Polynomial approximation of the smooth functional representation finds a spline approximation of the evaluated function to compute linearly independent basis for the gradient matrix. Finally, a linear model is used to compute the null space of the gradient matrix which indicates the coefficients of the underlying governing equation.

\begin{figure}[htbp]
    \centering

    \begin{subfigure}[t]{0.48\linewidth}
        \centering
        \includegraphics[width=\linewidth]{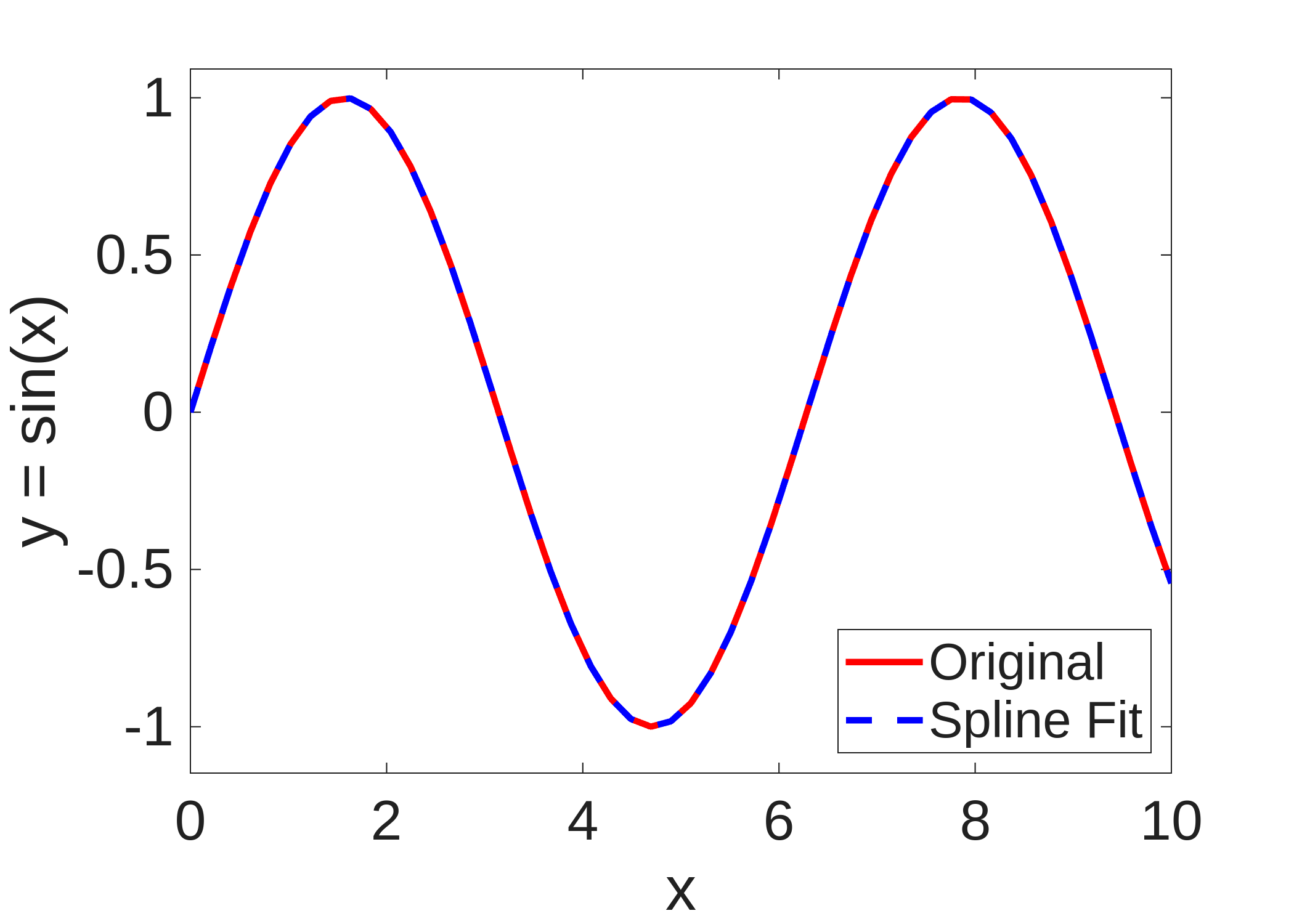}
        \caption{Without noise}
        \label{fig:spline_fitting_denoise}
    \end{subfigure}
    \hfill
    \begin{subfigure}[t]{0.48\linewidth}
        \centering
        \includegraphics[width=\linewidth]{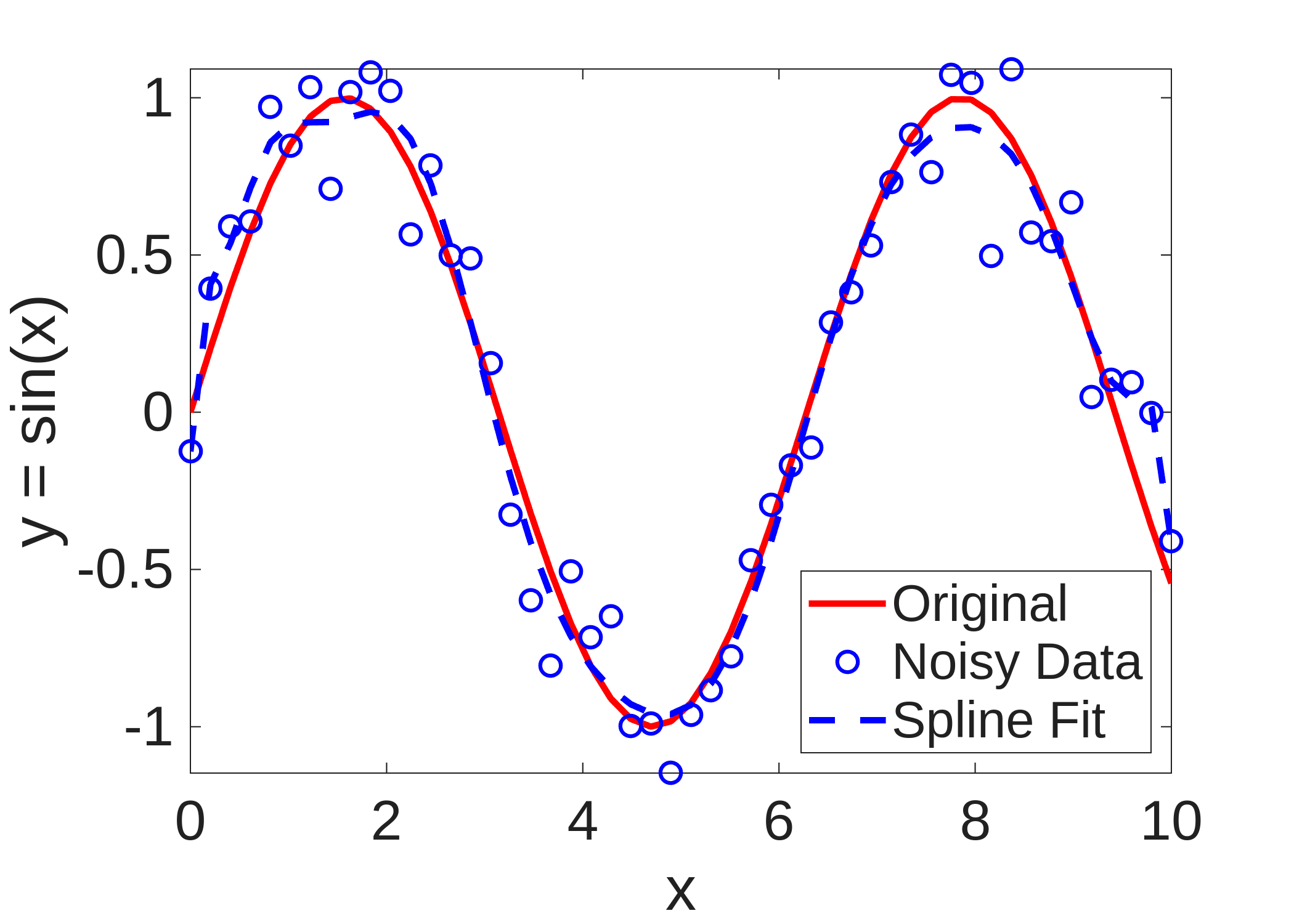}
        \caption{With noise}
        \label{fig:spline_fitting_noise}
    \end{subfigure}
    
    \caption{Comparison of spline fitting with and without noise.}
    \label{fig:spline_fitting_all}
\end{figure}

While the approximate general solution step allows to obtain smooth derivative information directly from the approximated function, for many functions, we may not be able to obtain linearly independent derivative information.
For linear model such as Eq. \ref{sec:methodology:eq:general_ODE}, we need linearly independent basis in the gradient matrix. Derivatives of nonlinear functions, such as trigonometric and exponential, are linearly dependent on the function itself. So, using gradients information from the approximate general solution may not lead to an effective gradient matrix computation. On the other hand, derivatives of polynomial functions are always linearly independent on each other which can be shown using the Wronskian determinant method \cite{voorhoeve1975wronskian}. However, directly obtaining a polynomial approximation using spline method is susceptible to noise present in the data. Figure \ref{fig:spline_fitting_all} highlights this issue by comparing spline fitting for smooth and noisy data. It shows the spline fitting for a simple function: $y = sin(x)$. We observe that, in the presence of noise, the spline does not fit the data effectively. This becomes an issue while computing the gradient matrix. If the gradient matrix has incorrect gradient information, then the linear model is not able to discover the proper coefficients of the physical system. That is why, we perform smooth approximate general solution followed by polynomial approximation.


\subsection{Approximate General Solution}

For a given coefficient set $\mathbf{\hat{C}} (\mathbf{\hat{C}}: {\hat{C}_{0}, \hat{C}_{1}, \cdots, \hat{C}_{P}})$,
it is known that Eq. \ref{sec:methodology:eq:general_ODE} takes a solution of the form:

\begin{ceqn}
    \begin{equation}
\label{sec:methodology:eq:analytical_form}
    y = A \exp(\lambda x)
\end{equation}
\end{ceqn}

\noindent where $\lambda$ are the eigenvalues that satisfy the characteristic equation

\begin{ceqn}
    \begin{equation}
\label{sec:methodology:eq:polynomial_form}
    \sum_{p=0}^{P}\hat{C}_{p}\lambda^p = 0.
\end{equation}
\end{ceqn}

\noindent To determine $\lambda$ we first construct the companion matrix $\mathbf{M} \in \mathbb{R}^{P \times P}$ given by
\begin{ceqn}
\begin{equation}
\label{matrix:companion_matrix}
\mathbf{M} = \begin{bmatrix}
0 & 1 & 0 & \cdots & 0 \\
0 & 0 & 1 & \cdots & 0 \\
\vdots & \vdots & \vdots & \cdots & \vdots \\
\frac{-\hat{C}_{0}}{\hat{C}_{P}} & \frac{-\hat{C}_{1}}{\hat{C}_{P}} & \frac{-\hat{C}_{2}}{\hat{C}_{P}} & \cdots & 0 \\
\end{bmatrix}
\end{equation}
\end{ceqn}

\noindent and then determine the eigenvalues of the companion matrix by solving

\begin{ceqn}
    \begin{equation}
\label{sec:methodology_eq:methodology_determinant_form}
    det(\mathbf{M}-\lambda \mathbf{I}) = 0.
\end{equation}
\end{ceqn}

\noindent If the coefficient set $\mathbf{\hat{C}}$ is non-zero then the matrix $\mathbf{M}$ is well defined and Eq. \ref{sec:methodology_eq:methodology_determinant_form} is solvable since all rows are linearly independent. We have to note that there can be multiplicities among the eigenvalues (for example, double roots and triple roots). There are $Q$ ($Q \leq P$) numbers of unique eigenvalues. When $Q = P$, it indicates all the eigenvalues to be unique. Also, a unique eigenvalue $\lambda_{q} (q=0, 2, \cdots, Q)$ can have both real and imaginary components. With this information, we can compute $\alpha_{q}, \beta_{q}, \gamma_{q}$, which are the multiplicity, real component and imaginary component of a unique eigenvalue $\lambda_{q}$.
Using these parameter values we can obtain the eigenfunction corresponding to each eigenvalue as follows,


\begin{ceqn}
    \begin{equation}
\label{sec:methodology:eq:eigen_function_basis_value}
    E_{q} = \left( \sum_{j=0}^{\alpha_{q}}x^{j} \right) e^{(\beta_{q}x)}(cos(\gamma_{q}x) + sin(\gamma_{q}x))
\end{equation}
\end{ceqn}

\noindent where . The general functional solution of the problem is given by,


\begin{ceqn}
    \begin{equation}
\label{sec:methodology:eq:solution_general_ODE}
    y = \sum_{q = 0}^{Q} D_{q} E_{q} = \sum_{q = 0}^{Q} D_{q} \left( \sum_{j=0}^{\alpha_{q}}x^{j} \right) e^{(\beta_{q}x)}(cos(\gamma_{q}x) + sin(\gamma_{q}x))
\end{equation}
\end{ceqn}

\noindent where $D_{q}$ are the coefficients associated with each unique eigenfunction. We approximate $D_{q}$ by regression using the input-output $(\mathit{\hat{x}_{k}}, \mathit{\hat{y}_{k}})$ data, where $k = 1, 2, \cdots, m$ represents the data points. Specifically, we use the linear equation given by 

\begin{ceqn}
    \begin{equation}
\mathbf{\hat{E}}^\mathbf{T}\mathbf{D} = \mathbf{\hat{y}}
\label{sec:methodology:eq:GDY}
\end{equation}
\end{ceqn}

\noindent where:
\begin{itemize}
    \item \( \mathbf{D} = [D_0, D_1, \cdots, D_{q}, \cdots, D_{Q}]^\mathrm{T} \in \mathbb{R}^{Q+1} \) is the coefficient vector of the eigenfunction,
    \item \( \mathbf{\hat{y}} = [\hat{y}_1, \hat{y}_2, \cdots, \hat{y}_{k}, \cdots, \hat{y}_m]^\mathrm{T} \in \mathbb{R}^{m} \) is the target output vector and
    \item \( \mathbf{\hat{E}} \in \mathbb{R}^{Q \times m} \) is the eigenfunction basis matrix with entries \( E_{q,k} \) computed through Eqs. \ref{matrix:companion_matrix}  - \ref{sec:methodology:eq:eigen_function_basis_value} for a fixed set of $\mathbf{\hat{C}}$.
\end{itemize}

\noindent Since $\mathbf{\hat{E}}$ is a rectangular system with $m > Q$ (overdefined system), we use Moore-Penrose pseudoinverse of $\mathbf{\hat{E}}$ to compute $\mathbf{D}$ given as follows

\begin{ceqn}
    \begin{equation}
\mathbf{D} = (\mathbf{\hat{E}}\mathbf{\hat{E}}^\mathbf{T})^{-1}\mathbf{\hat{E}}\mathbf{\hat{y}}.
\label{sec:methodology:eq:D_solution}
\end{equation}
\end{ceqn}

\noindent Eq. \ref{sec:methodology:eq:D_solution} gives D as the ``best least square" solution to the linear system in Eq. \ref{sec:methodology:eq:GDY}.
We compute $\mathbf{\hat{y}}_{\textbf{pred}}$ using Eq. \ref{sec:methodology:eq:solution_general_ODE} which may not be same as $\mathbf{\hat{y}}$ due to pseudoinverse calculation. The loss function to estimate the error is given by

\begin{ceqn}
\begin{equation}
\label{sec:methodology:eq:mse}
\mathrm{L}(\mathbf{\hat{y}}_{\textbf{pred}}, \mathbf{\hat{y}}) = \frac{1}{m} \sum_{k=1}^{m} \left( \hat{y}_{\text{pred},k} - \hat{y}_k \right)^2
\end{equation}
\end{ceqn}

\begin{algorithm}
\caption{Genetic Algorithm (GA) for Optimal Coefficient Estimation}
\label{algorithm:GA}
\begin{algorithmic}[1]
\State \textbf{Input:} Data \((\hat{\mathbf{x}}, \hat{\mathbf{y}})\), population size \(N_p\), max generations \(T\), mutation rate \(r_m\), crossover rate \(r_c\), tolerance \(\epsilon\)
\vspace{0.5em}
\State \textbf{Initialize:}
\State \hspace{1em} Generate initial population of \(N_p\) real-valued vectors \(\hat{\mathbf{C}}^{(i)} \sim \mathcal{U}(\text{lb}, \text{ub})\)
\State \hspace{1em} Set default GA options:
\State \hspace{2em} \textbullet\ Crossover: \textit{intermediate} \cite{eshelman1993real}
\State \hspace{2em} \textbullet\ Mutation: \textit{adaptive feasible} \cite{deb2000efficient}
\State \hspace{2em} \textbullet\ Selection: \textit{stochastic uniform} \cite{goldberg1989genetic}
\State \hspace{2em} \textbullet\ Elitism: keep top \(5\%\) individuals
\vspace{0.5em}
\For{generation \(t = 1\) to \(T\)}
    \For{each individual \(\hat{\mathbf{C}}^{(i)}\)}
        \State Compute Fitness: \(L^{(i)} = \text{FitFunc}(\hat{\mathbf{x}}, \hat{\mathbf{y}}, \hat{\mathbf{C}}^{(i)})\)
    \EndFor
    \State Find best loss: \(L_{\text{min}} = \min_i L^{(i)}\)
    \If{\(L_{\text{min}} < \epsilon\)}
        \State \textbf{Break:} tolerance satisfied
    \EndIf
    \State Select parent pool using \textit{stochastic uniform selection}
    \State Apply \textit{intermediate crossover} with rate \(r_c\)
    \State Apply \textit{adaptive feasible mutation} with rate \(r_m\)
    \State Preserve elite individuals for next generation
\EndFor
\State \Return best-performing \(\hat{\mathbf{C}}^\star = \arg\min_{\hat{\mathbf{C}}^{(i)}} L^{(i)}\)
\end{algorithmic}
\end{algorithm}

\begin{algorithm}
\caption{Fitness Function of Genetic algorithm (FitFunc)}
\label{algorithm:FitFunc}
\begin{algorithmic}[1]
\State \textbf{Input:} Input data $\mathbf{\hat{x}}$, output data $\mathbf{\hat{y}}$, coefficient set $\mathbf{\hat{C}}$
\State Construct characteristic polynomial from $\mathbf{\hat{C}}$ using Eq.~\ref{sec:methodology:eq:polynomial_form}
\State Construct companion matrix $\mathbf{M}$ using Eq.~\ref{matrix:companion_matrix}
\State Compute eigenvalues: $\lambda \gets \texttt{eig}(\mathbf{M})$
\State Extract unique eigenvalues: $S' \gets \texttt{unique}(\lambda)$
\For{each $\lambda_q$ in $S'$}
    \State $\alpha_q \gets$ multiplicity of $\lambda_q$ in $\lambda$
    \State $\beta_q \gets \texttt{Re}(\lambda_q)$
    \State $\gamma_q \gets \texttt{Im}(\lambda_q)$
\EndFor
\For{each $x_k$ in $\mathbf{\hat{x}}$}
    \For{each $q = 0, \dots, Q$}
        \State Compute $E_{q,k}$ using Eq.~\ref{sec:methodology:eq:eigen_function_basis_value}
    \EndFor
\EndFor
\State Assemble matrix $\mathbf{\hat{E}}$ from $E_{q,k}$
\State Solve for $\mathbf{D}$ using Eq.~\ref{sec:methodology:eq:D_solution}
\State Compute prediction $\mathbf{\hat{y}_{pred}}$ using Eq.~\ref{sec:methodology:eq:GDY}
\State Compute Loss $\mathrm{L}$ using Eq. ~\ref{sec:methodology:eq:mse}
\State \Return Loss $\mathrm{L}$
\end{algorithmic}
\end{algorithm}

We use data-driven optimization algorithms such as the genetic algorithm \cite{mirjalili2019genetic} to find the optimal set of coefficients $\mathbf{\hat{C}}$. The genetic algorithm is given in Algorithm \ref{algorithm:GA}. We use a custom fitness function for the genetic algorithm given in Algorithm \ref{algorithm:FitFunc}.
\noindent Now, a question arises from the discussion: Since we find the optimal parameter set $\mathbf{\hat{C}}$ from the data-driven optimization model, why is further analysis necessary? Why aren't the coefficient set $\mathbf{\hat{C}}$ interpreted as the ODE coefficients? 
Our objective is to obtain a high-fidelity smooth approximate general solution of the noisy data. So, we have considered a higher order ODE formulation with high degree of freedom to make the search flexible. Also, the data observed are in specific bounds of time. There can be multiple ODEs that have similar response in that specific input domain. Our objective here is not to immediately find the ODE where the data are coming from but a functional form which smoothly approximates the data. For this reason, we empirically observe that the optimized coefficient set is not representative of the underlying ODE. To arrive at the final values of the coefficient $C$, we perform the operations described in the subsequent subsections \ref{subsec:spline_approximation} and \ref{subusec:linear_modeling}.


\subsection{Polynomial Approximation using Spline}
\label{subsec:spline_approximation}
After the evaluation of parameters of Eq. \ref{sec:methodology:eq:solution_general_ODE} fitting the data, a polynomial approximation of the function in Eq. \ref{sec:methodology:eq:solution_general_ODE} is performed. The primary reason for this is to find a functional form where the derivatives of the function is linearly independent of each other. Although the functional form in Eq. \ref{sec:methodology:eq:solution_general_ODE} represents the general solution of the ODE in Eq. \ref{sec:methodology:eq:general_ODE}, it may not have linearly independent derivatives. 
For example, consider an illustrative case where the ODE has a functional solution $y = e^{2x}$ due to $\alpha$, $\beta$ and $\gamma$ taking values of $0, 2, 0$ respectively. The first derivative of the function will be $y' = 2e^{2x}$ which is linearly dependent on $y$. Therefore, a linear model will fail to determine the coefficients $C$ in this situation.
Polynomials are, on the other hand, a set of functions which have linearly independent gradient properties. Since our objective is to discover an ODE from a set of data, we consider the general order of the ODE to be of order $P$, which can be a high value. For this reason, a higher order polynomial approximation is suuitable. 

\textbf{Initial spline fit:} Let $\mathit{t}$ be a non-decreasing sequence of knots selected uniformly over the domain of inputs ($\hat{x}$) such that

\begin{ceqn}
    \begin{equation}
    \label{eq:knot_vector_initialization}
        t_{0} \leq t_{1} \leq t_{2} \leq \cdots \leq t_{P+\omega}.
    \end{equation}
\end{ceqn}

\noindent where $P$ is the number of basis functions (here order of ODE), $\omega$ is the order of the B-spline (here $\omega=P$). At a given knot $t_{s}$, the $s^{th}$ B-Spline function $B_{s}^{(\omega)}$ of order $\omega$ is given by 

\begin{ceqn}
\begin{equation}
\label{eq:bspline_basis_definition}
B_s^{(\omega)}(x) = 
\begin{cases}
1, & \text{if } \omega = 1 \text{ and } t_s \le x < t_{s+1}, \\
0, & \text{if } \omega = 1 \text{ and otherwise}, \\
\frac{x - t_s}{t_{s+\omega-1} - t_s} B_s^{(\omega - 1)}(x) + 
\frac{t_{s+\omega} - x}{t_{s+\omega} - t_{s+1}} B_{s+1}^{(\omega - 1)}(x), 
& \text{if } \omega > 1.
\end{cases}
\end{equation}
\end{ceqn}

\noindent For $\omega > 1$, $B_{s}^{(\omega)}$ is computed recursively as the weighted sum of two lower-order ($\omega-1$) basis functions, $B_{s}^{(\omega-1)}$ and $B_{s+1}^{(\omega-1)}$. Given the approximate general solution of the data, initial spline approximation of the function is given by

\begin{ceqn}
    \begin{equation}
\label{sec:methodology:eq:spline-approximation}
    S(x) = \sum_{s=0}^{P} \mu_{s}B_{s}^{(\omega)} (x)
\end{equation}
\end{ceqn}

\noindent where $\mu_{s}$ is the unknown coefficient of $B_{s}^{(\omega)} (x)$. To determine the coefficient $\mu_{s}$, we solve the following optimization problem:

\begin{ceqn}
\begin{equation}
\begin{aligned}
\min_{\mu_{s}} \quad & \sum_{k=1}^{m} \left( y_{pred,k} - S(\hat{x}_{k}) \right)^{2} \\
\text{s.t.} \quad & S(\hat{x}_{k}) = \sum_{s=0}^{P} \mu_{s} B_j^{(\omega)}(\hat{x}_{k}).
\end{aligned}
\label{eq:spline_fit_optimization}
\end{equation}
\end{ceqn}

\noindent Since $\hat{x}_{k}$ values are known, we can precompute the basis functions $B_{s}^{(\omega)} (\hat{x})$. Therefore, the optimization problem becomes a linear constrained quadratic objective system which has an analytical solution.

\textbf{Adaptive knot selection:} Following the initial spline computation, adaptive knot selection is performed. At a data point $\hat{x_k}$, the approximation error between the initial spline fit $S(\hat{x}_{k})$ and the functional value $y_{pred,k}$ is given by

\begin{ceqn}
    \begin{equation}
    \label{eq:adap_knot1}
        r_{k} = y_{pred,k} - S(\hat{x}_{k}).
    \end{equation}
\end{ceqn}

\noindent We define a piecewise error function ($\phi$) over subintervals of ($\hat{x}_{1}, \hat{x}_{2}, \cdots, \hat{x}_{m}$) for each interval between knots $[t_{j}, t_{j+1}]$ as follows:

\begin{ceqn}
    \begin{equation}
    \label{eq:adap_knot2}
        \phi(t_{j}) = \sum_{x_{k} \in [t_{j}, t_{j+1}]} r_{k}^{2}.
    \end{equation}
\end{ceqn}

\noindent Following the error calculation, a refinement strategy is implemented based on a threshold value ($\tau$) of error

\begin{ceqn}
\begin{equation}
j \in \mathcal{J} \quad \text{if} \quad \phi(t_j) > \tau, \quad \forall j \in [0, P + \omega]
\end{equation}
\end{ceqn}

\begin{algorithm}
\caption{Spline-Based Polynomial Approximation with Adaptive Knot Refinement}
\label{algorithm:spline}
\begin{algorithmic}[1]
\State \textbf{Input:} Data $\{(\hat{x}_k, y_{\text{pred},k})\}_{k=1}^m$, spline order $\omega = P$, threshold $\tau$

\vspace{0.2cm}
\State \textbf{Initialization:}
\State \hspace{0.4cm} Generate uniformly spaced knot vector $t$ as described in Eq.~\eqref{eq:knot_vector_initialization}
\State \hspace{0.4cm} Compute B-spline basis functions $B_s^{(\omega)}(x)$ using Eq.~\eqref{eq:bspline_basis_definition}

\vspace{0.2cm}
\State \textbf{Step 1: Initial Spline Fit}
\State Fit initial spline $S(x)$ by solving the optimization problem in Eq.~\eqref{eq:spline_fit_optimization}

\vspace{0.2cm}
\State \textbf{Step 2: Adaptive Knot Refinement}
\Repeat
    \State Compute residuals $r_k$ as defined in Eq.~\eqref{eq:adap_knot1}
    \State Compute piecewise error $\phi(t_j)$ on each interval using Eq.~\eqref{eq:adap_knot2}
    \If{$\phi(t_j) > \tau$ for any $j \in [0, P + \omega]$ (see Eq.~\eqref{eq:adap_knot2})}
        \State Identify $\mathcal{J} = \{ j \mid \phi(t_j) > \tau \}$ (see Eq.~\eqref{eq:adap_knot2})
        \State Insert new knots $\tilde{t}_j \forall j \in \mathcal{J}$ (see Eq.~\eqref{eq:adap_knot3})
        \State Update knot vector $t$ (see Eq.~\eqref{eq:adap_knot4})
    \EndIf
    \State Recompute basis functions and refit the spline using Eq.~\eqref{eq:spline_fit_optimization}
\Until{No new knots satisfy the refinement condition}
\State \Return Final high-fidelity spline $S(x)$
\end{algorithmic}
\end{algorithm}

\noindent to select intervals with high error values. Since, we want to obtain a high fidelity spline approximation of the functional data, $\tau$ is set to $10^{-6}$. Using $\tau$, we create a new set of knots $\mathcal{J}$ which requires refinement. For each $j \in \mathcal{J}$, we insert a new knot $\tilde{t}_{j}$ given by

\begin{ceqn}
    \begin{equation}
    \label{eq:adap_knot3}
        \tilde{t}_{j} = \frac{t_{j} + t_{j+1}}{2}
    \end{equation}
\end{ceqn}

\noindent and augment the present knot vector using

\begin{ceqn}
    \begin{equation}
    \label{eq:adap_knot4}
        t := t \cup \{\tilde{t}_{j}:j \in \mathcal{J}\}.
    \end{equation}
\end{ceqn}

\noindent The new knot vector $t$ is used to perform the calculation from Eq. \ref{eq:bspline_basis_definition} to \ref{eq:adap_knot4} until convergence is reached for all knots in the knot vector $t$. The algorithm for the spline calculation is provided in Algorithm \ref{algorithm:spline}.

Following the spline approximation $S(x)$ of the approximate general solution of the original data, we compute the gradient basis function of order $p (={0, 1, \cdots, P})$ using the following equation:

\begin{ceqn}
    \begin{equation}
    \label{eq:spline_gradient_relation}
    \frac{d^{p}y}{dx^{p}} =
    \begin{cases}
        S(x), & \text{if } p = 0, \\
        \frac{d^{p} S(x)}{dx^{p}}, & \text{if } p \geq 1
        \end{cases}
    \end{equation}
\end{ceqn}

\subsection{Linear Coefficient Estimation}
\label{subusec:linear_modeling}
Let $y(x)$ represents the function describing the solution of the ODE in Eq. \ref{sec:methodology:eq:general_ODE}. Then the following holds:

\begin{ceqn}
    \begin{equation}
\label{eq:general_ODE_y-x}
    \sum_{p=0}^{P} C_{p} \frac{d^{p}y_{k}}{dx_{k}^{p}} = 0, k=1, 2, \cdots, m.
\end{equation}
\end{ceqn}

\noindent where $\frac{d^{p}y_{k}}{dx_{k}^{p}}$ are computed using Eq. $\ref{eq:spline_gradient_relation}$. If $\mathbf{G}$ is the gradient matrix of y such that its element $G_{p, k} = \frac{d^{p}y_{k}}{dx_{k}^{p}}$, then we obtain a matrix system of equations given by

\begin{ceqn}
    \begin{equation}
\label{eq:linear_eq_c_G_matrix}
    \mathbf{G}^{T}\mathbf{C} = 0
\end{equation}
\end{ceqn}

\noindent where $\mathbf{G} \in \mathbb{R}^{(P+1, m)}$ and $\mathbf{C} \in \mathbb{R}^{(P+1)}$. Since the spline $S$ is a polynomial of order $P$, it has $P+1$ linearly independent basis $(x^{0}, x^{1}, \cdots, x^{P})$. Therefore, its gradient matrix $G$ has $\leq P+1$ linearly independent rows. Therefore, we can find a reduced matrix $\tilde{\mathbf{G}}$ given by,

\begin{ceqn}
    \begin{equation}
\label{eq:linear_model_system}
    \mathbf{\tilde{G}}^{T}\mathbf{C} = 0
\end{equation}
\end{ceqn}

\noindent where $\mathbf{\tilde{G}} \in \mathbb{R}^{(P+1, L)}$ $(L \leq (P+1))$. For a non-trivial solution of $C$, $\mathbf{\tilde{G}}$ must be a rank-deficient matrix. Therefore, $L < (P+1)$. Since the data are assumed to follow a unique governing equation, there is a unique non-trivial coefficient set $\mathbf{C}$. So, the question becomes how we find $\mathbf{\tilde{G}}$ and how we solve Eq. \ref{eq:linear_model_system}
 to find the coefficient set $\mathbf{C}$ over the null space of $\mathbf{\tilde{G}}$ (see Eq. \ref{eq:linear_model_system}). To this end, we use singular value decomposition (SVD) to find the null space of $\mathbf{\tilde{G}}$. The SVD of $\mathbf{\tilde{G}}$ is given by

\begin{ceqn}
     \begin{equation}
     \mathbf{\tilde{G}} = \mathbf{U} \mathbf{\Sigma} \mathbf{V}^\top,
 \end{equation}
\end{ceqn}

\noindent where:
 
 \begin{itemize}
     \item $ \mathbf{U} \in \mathbb{R}^{(L) \times (L)} $ is an orthogonal matrix.
     \item $ \mathbf{\Sigma} \in \mathbb{R}^{(L) \times (P+1)} $ is a diagonal matrix with non-negative singular values $ \sigma_1 \geq \sigma_2 \geq \cdots \geq \sigma_{L} $, and
     \item $ \mathbf{V} \in \mathbb{R}^{(P+1) \times (P+1)} $ is an orthogonal matrix.
 \end{itemize}

\noindent The null space of $\mathbf{\tilde{G}}$ is given by subset of columns of the right singular matrix $\mathbf{V}$, which corresponds to zero singular values in $\mathbf{\Sigma}$. 
If $\mathrm{rank}(\mathbf{\tilde{G}}) = L < (P+1)$, then we have a non-trivial null space (all elements of null space are not $0$) of the matrix $\mathbf{\tilde{G}}$. The number of unique null space basis vectors is given by

\begin{ceqn}
     \begin{equation}
     \Gamma = (P+1) - L.
 \end{equation}
\end{ceqn}

\noindent For any rank-deficient matrix, there will be $\Gamma$ number of non-trivial null space basis vectors. These basis vectors are present in the last $\Gamma$ columns of the right singular matrix $\mathbf{V}$. We assume that the data are generated from a unique governing equation $\Gamma = 1$ which corresponds to only one unique null space. In practice, if $\Gamma > 1$, then the last column of the null space basis vector is considered which corresponds to the ``least squares solution with minimum Euclidean norm" \cite{trefethen1997numerical}. 
With these the coefficient vector, $\mathbf{C}$, of the ODE is obtained by taking the last column of the right orthogonal matrix $\mathbf{V}$:
\begin{ceqn}
    \begin{equation}
    \mathbf{C} = \mathbf{V}_{(:,P+1)}
    \end{equation}
\end{ceqn}

Several questions may arise from the discussion. How many data points ($m$) should we choose for the matrix $\mathbf{G}$? Since $\mathbf{G}$ represents an overdefined system which is reduced to a smaller rank-deficient system $\mathbf{\tilde{G}}$, will choosing $L$ points for $\mathbf{G}$ be enough to get the optimum null space? Here, if we choose $L$ points to create the matrix $\mathbf{G}$, then there is a possibility that some of the rows of $\mathbf{G}$ are linearly dependent on other rows, giving an incomplete rank-deficient matrix $\mathbf{\hat{G}}$ where $\mathrm{rank}(\mathbf{\hat{G}}) < \mathrm{rank}(\mathbf{\tilde{G}})$. So, to obtain a complete rank-deficient matrix, we need to sample enough points from the spline approximation. Here, in our study, we have sampled data points of order $10^{3}$ to create the gradient matrix $\mathbf{G}$.

\section{Results and Discussion}
\label{sec:results}

We have performed two case studies to demonstrate the overall approach for data-driven ODE discovery. First, we consider a physical spring mass system as a benchmark example of harmonic systems with well-defined dynamics. 
We present the model's performance for the spring mass system, model's capability to discover hidden properties of the system from the observed data, and dependence of different parameter and hyperparameters on the model's performance.
The second case study was performed for a chemical engineering problem. Here, we consider photolytic degradation for a multicomponent mixture of estrogen disrupting chemicals (EDCs) that empirically follows first-order degradation kinetics. 

\subsection{Case Study 1: Spring-Mass System}
\label{subsec:results&discussions}
\begin{figure}[htbp]
    \centering
    \includegraphics[width=0.8\linewidth]{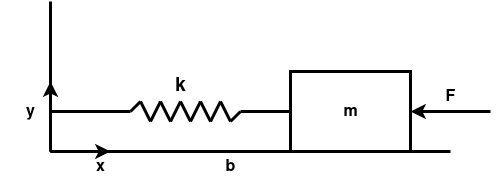}
    \caption{Spring Mass system.}
    \label{fig:spring_mass}
\end{figure}

The spring mass system is a classical example of a harmonic oscillator system which follows second-order kinetics.
There have been theoretical approaches \citep{nylen1997inverse, alvares2024discovering} for modeling and simulation of a spring mass system. Here, for the spring problem, we consider that a block is attached to a spring and pushed along the negative x direction (See Figure \ref{fig:spring_mass}). The mass of the block, friction coefficient of the surface and spring constant of the spring are denoted by m, b, k respectively. 

\begin{table}[]
\centering
\caption{Dataset Configuration for case studies.}
\begin{tabular}{lrrr}
\hline
Parameters            & \multicolumn{1}{l}{Underdamped} & \multicolumn{1}{l}{Critical Damped} & \multicolumn{1}{l}{Overdamped} \\ \hline
Mass (m)              & 4                               & 1                                   & 2                              \\
Damping Factor (b)    & 2                               & 2                                   & 4                              \\
Spring Constant (k)   & 1                               & 1                                   & 1                              \\
Initial Position (x0) & 0.4                             & 0.4                                 & 0.4                            \\
Initial Velocity (x1) & -0.6                            & -0.6                                & -0.6                           \\
Time Duration (T)     & 20                              & 20                                  & 20                             \\
Data Points           & 1000                            & 1000                                & 1000                           \\
Gaussian Noise Mean   & 0.5                             & 0.5                                 & 0.5                            \\
Gaussian Noise SD     & 0.1                             & 0.1                                 & 0.1                            \\
Gaussian Noise Scale  & 0.001                           & 0.001                               & 0.001                          \\ \hline
\end{tabular}
\label{tab:dataset_config_case_studies}
\end{table}

The dataset used for analysis is synthetically generated using an analytical solution of a spring mass system under different scenarios (see Appendix \ref{appx:spring_mass_data_generation}).
We performed analysis on the spring mass system for both noisy data and data without noise. The configuration for generating the datasets for all three is provided in Table \ref{tab:dataset_config_case_studies}. Parameters other than mass and damping factor are considered identical for all cases. The spring constant $(k)$ is considered to be $1$. Multiplying the predicted governing system with any scaling factor will not change the response of the system since it is a homogeneous constant coefficient system. Gaussian noise is added to the data with defined properties (mean, standard deviation, and scale). We have considered a noise scale of $0.001$. The negative initial velocity indicates that the displacement is initially in the negative $x$ direction. So, a pushing force is applied to the mass. A time duration of $20$ s is considered.

Table \ref{tab:genetic_algorithm_configuration} shows the configuration of the genetic algorithm-based optimizer. During experimentation, we observed that model convergence is faster for large population size. To improve the fidelity of the model, a very low functional tolerance is considered. Additionally, a bound is provided for all the parameters that are estimated. Keeping the search space small through tight bounds increases the speed of convergence of the model but reduces the generalization capability of the model. Also, the parameters being estimated from the model may not align with the true solution. Since the objective of the algorithm is to determine a smooth functional form from the noisy data, the bounds are found to be useful.

\begin{table}[]
\centering
\caption{Genetic Algorithm Configuration for case studies.}
\begin{tabular}{lr}
\hline
Properties          & \multicolumn{1}{l}{Value} \\ \hline
Population Size     & 750                       \\
Maximum Generations & 300                       \\
Function Tolerance  & 1.00E-28                  \\
Lower Bound         & -10                       \\
Upper Bound         & 10                        \\ \hline
\end{tabular}
\label{tab:genetic_algorithm_configuration}
\end{table}

\begin{figure}[htbp]
    \centering

    \begin{subfigure}[t]{0.49\linewidth}
        \centering
        \includegraphics[width=\linewidth]{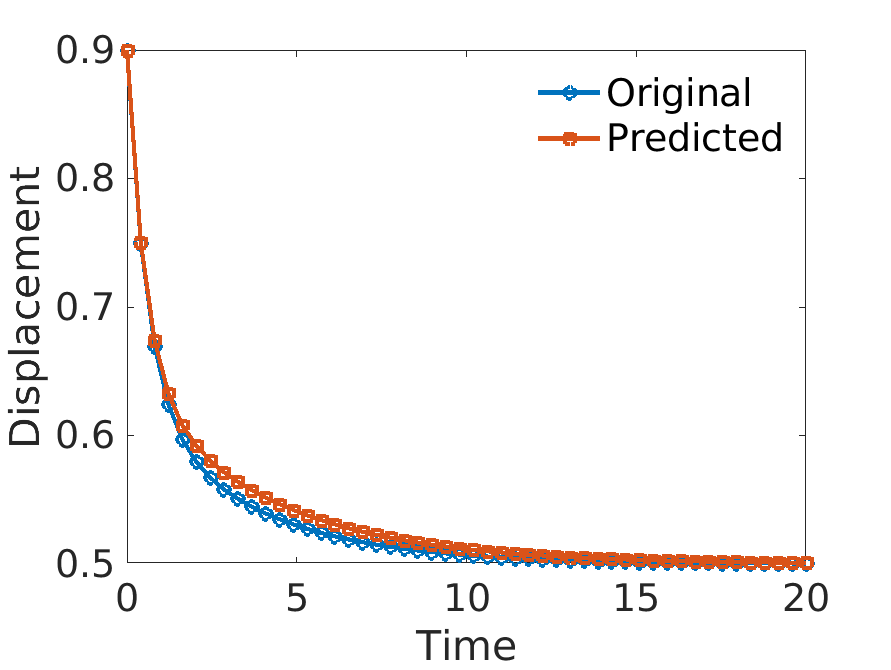}
        \caption{Overdamped system}
        \label{fig:spring_overdamped_condition}
    \end{subfigure}
    \hfill
    \begin{subfigure}[t]{0.49\linewidth}
        \centering
        \includegraphics[width=\linewidth]{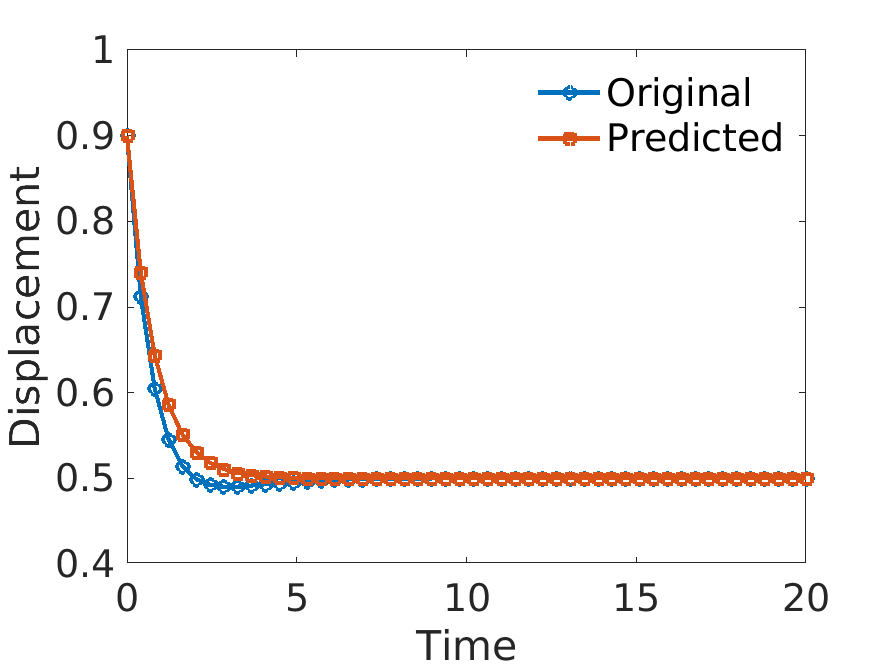}
        \caption{Critical damped system}
        \label{fig:spring_critical_damped_condition}
    \end{subfigure}
    \hfill
    \begin{subfigure}[t]{0.49\linewidth}
        \centering
        \includegraphics[width=\linewidth]{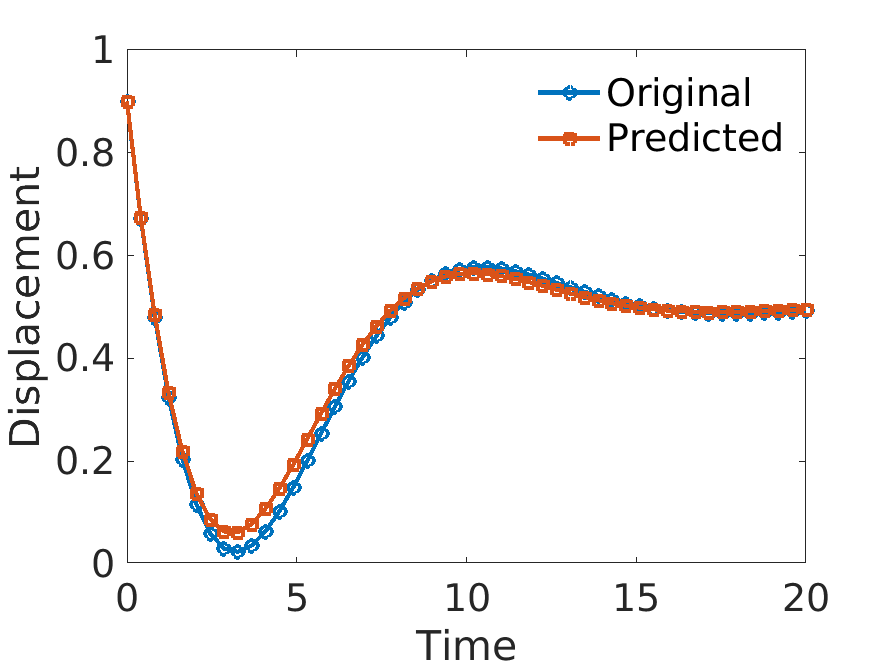}
        \caption{Underdamped system}
        \label{fig:spring_underdamped_condition}
    \end{subfigure}

    \caption{Comparison of model predictions for overdamped, critically damped, and underdamped spring-mass systems.}
    \label{fig:spring_damping_regimes}
\end{figure}


\begin{table}[htbp]
\centering
\caption{Comparison of Model Predictions With and Without Noise.}
\small
\begin{tabular}{llll}
\toprule
\textbf{Noise} & \textbf{Case} & \textbf{Original ODE} & \textbf{Predicted ODE} \\
\midrule

\multirow{3}{*}{With Noise} 
& Overdamped & $2\frac{d^2x}{dt^2} + 4\frac{dx}{dt} + 1.00x = 0$ & 
$2.004\frac{d^2x}{dt^2} + 4.001\frac{dx}{dt} + 1.00x = 0$ \\

& Critical Damped & $\frac{d^2x}{dt^2} + 2\frac{dx}{dt} + 1.00x = 0$ & 
$0.829\frac{d^2x}{dt^2} + 1.885\frac{dx}{dt} + 1.00x = 0$ \\

& Underdamped & $4\frac{d^2x}{dt^2} + 2\frac{dx}{dt} + 1.00x = 0$ & 
$4.507\frac{d^2x}{dt^2} + 2.139\frac{dx}{dt} + 1.00x = 0$ \\

\midrule

\multirow{3}{*}{Without Noise} 
& Overdamped & $2\frac{d^2x}{dt^2} + 4\frac{dx}{dt} + 1.00x = 0$ & 
$2.03\frac{d^2x}{dt^2} + 1.03\frac{dx}{dt} + 1.00x = 0$ \\

& Critical Damped & $\frac{d^2x}{dt^2} + 2\frac{dx}{dt} + 1.00x = 0$ & 
$1.02\frac{d^2x}{dt^2} + 2.01\frac{dx}{dt} + 1.00x = 0$ \\

& Underdamped & $4\frac{d^2x}{dt^2} + 2\frac{dx}{dt} + 1.00x = 0$ & 
$4.01\frac{d^2x}{dt^2} + 2.01\frac{dx}{dt} + 1.00x = 0$ \\

\bottomrule
\end{tabular}
\label{tab:spring_model_prediction}
\end{table}

Figure \ref{fig:spring_damping_regimes} shows an example of the on the spring mass system with three different responses. Figure \ref{fig:spring_overdamped_condition} shows the prediction on an overdamped system. From Table \ref{tab:spring_model_prediction}, we can compare the original ODE used to generate the data for overdamped system with noise and the predicted system providing smooth response. From the approximate general solution results for this case, we observe the mean square error (MSE) value between predicted functional form and the original data is $3.201 \times 10^{-7}$ which is considerably low. This suggests that the approximate general solution plays a useful role in the presence of noisy data. Interestingly, the coefficient estimated by the genetic optimizer during this approximation is $(0.0654, -3.4774, 7.5927, -0.7251, 7.9382, 9.9778)$ where the indices correspond to the order of the estimated ODE. Although the values of the coefficients do not correlate with the original ODE, the low error value mean the functional form of the ODE generated using this coefficient is able to find a high-fidelity smooth functional form of the governing behavior of the system from where the data was generated. Additionally, prediction results for critical damped case and underdamped case are shown in Figures \ref{fig:spring_critical_damped_condition} and \ref{fig:spring_underdamped_condition}, respectively. The MSE losses for approximate general  for both cases are $9 \times 10^{-9}$ and $1.2 \times 10^{-9}$, respectively, indicating the genetic algorithm is able to approximate smooth functional forms for all different responses of the spring mass system.

\begin{figure}[htbp]
    \centering

    \begin{subfigure}[t]{0.48\linewidth}
        \centering
        \includegraphics[width=\linewidth]{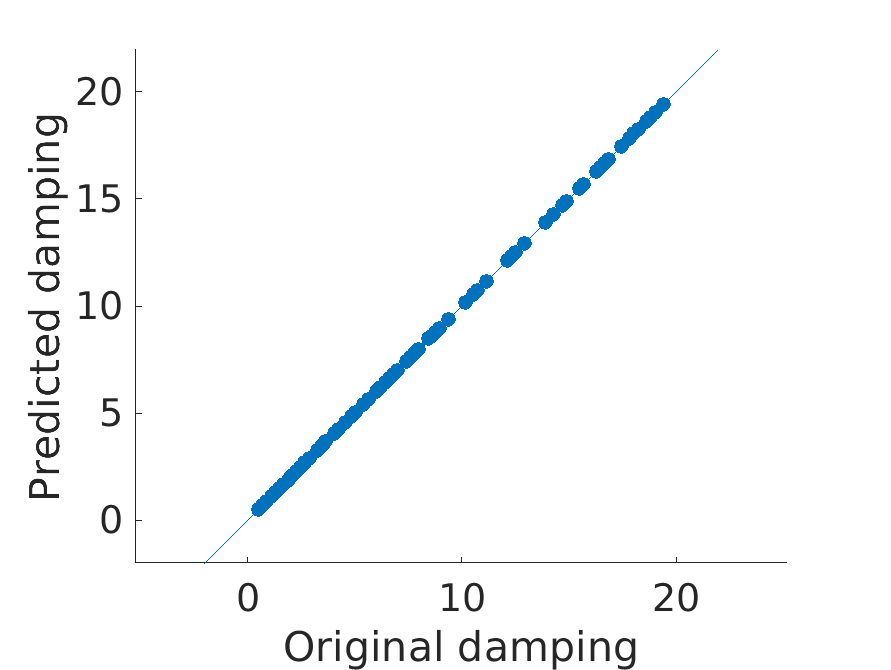}
        \caption{Damping factor}
        \label{fig:damping_reg_no_noise}
    \end{subfigure}
    \begin{subfigure}[t]{0.48\linewidth}
        \centering
        \includegraphics[width=\linewidth]{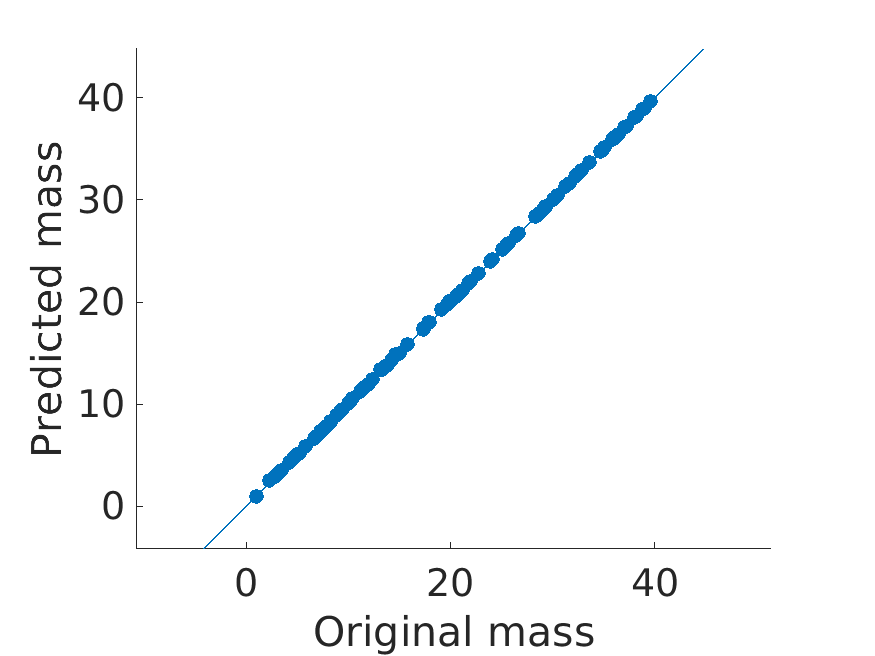}
        \caption{Mass}
        \label{fig:mass_reg_no_noise}
    \end{subfigure}
    \caption{Regression plots for damping factor and mass without noise.}
    \label{fig:reg_no_noise}
\end{figure}

\begin{figure}[htbp]
    \centering

    \begin{subfigure}[t]{0.325\linewidth}
        \centering
        \includegraphics[width=\linewidth]{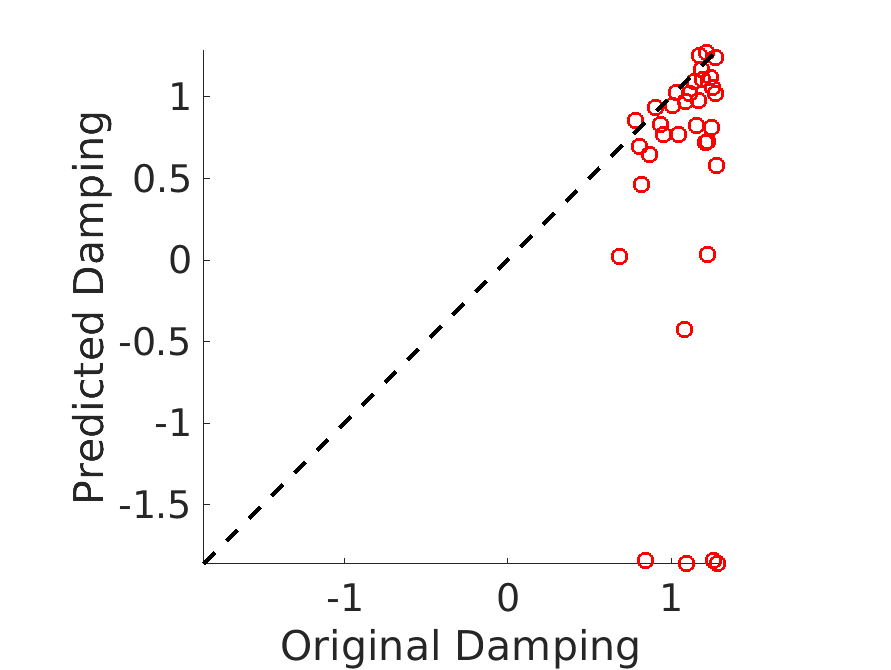}
        \caption{Overdamped system}
        \label{fig:spring_regression_comparison_damping_factor_overdamped_condition}
    \end{subfigure}
    \hfill
    \begin{subfigure}[t]{0.325\linewidth}
        \centering
        \includegraphics[width=\linewidth]{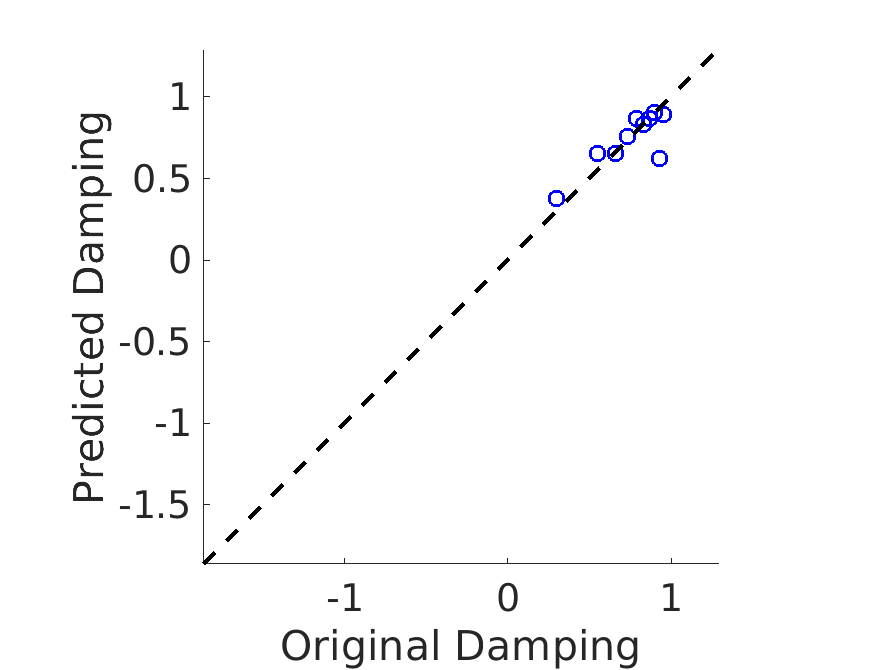}
        \caption{Critical damped system}
        \label{fig:spring_regression_comparison_damping_factor_critical_damped_condition}
    \end{subfigure}
    \hfill
    \begin{subfigure}[t]{0.325\linewidth}
        \centering
        \includegraphics[width=\linewidth]{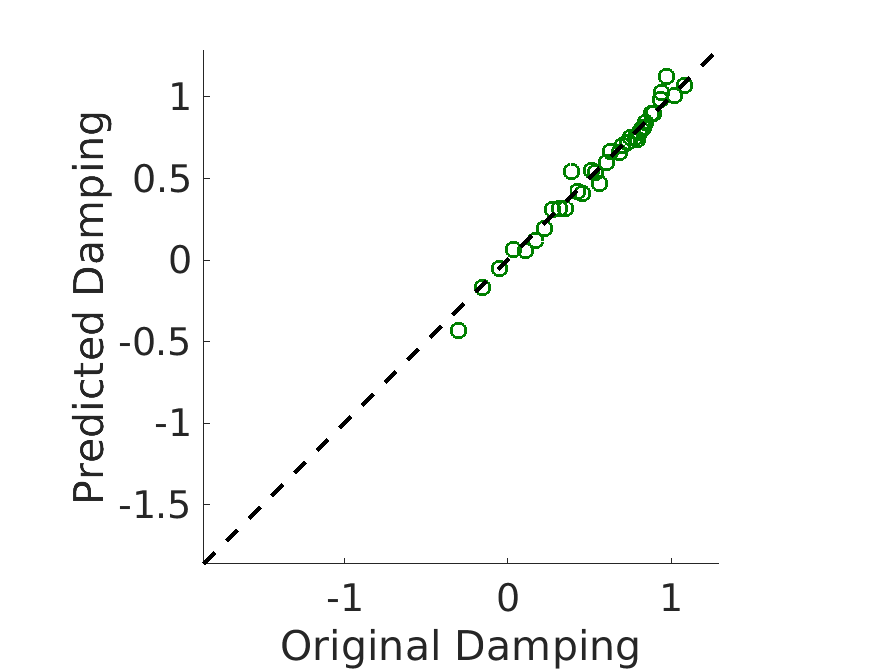}
        \caption{Underdamped system}
        \label{fig:spring_regression_comparison_damping_factor_underdamped_condition}
    \end{subfigure}

    \vspace{1em}

    \begin{subfigure}[t]{0.325\linewidth}
        \centering
        \includegraphics[width=\linewidth]{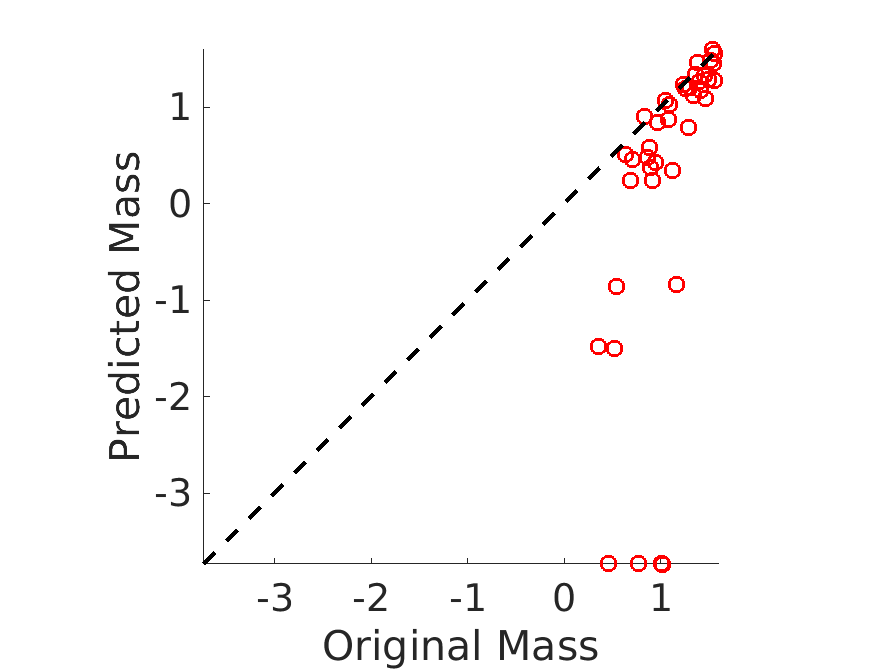}
        \caption{Overdamped system}
        \label{fig:spring_regression_comparison_mass_overdamped_condition}
    \end{subfigure}
    \hfill
    \begin{subfigure}[t]{0.325\linewidth}
        \centering
        \includegraphics[width=\linewidth]{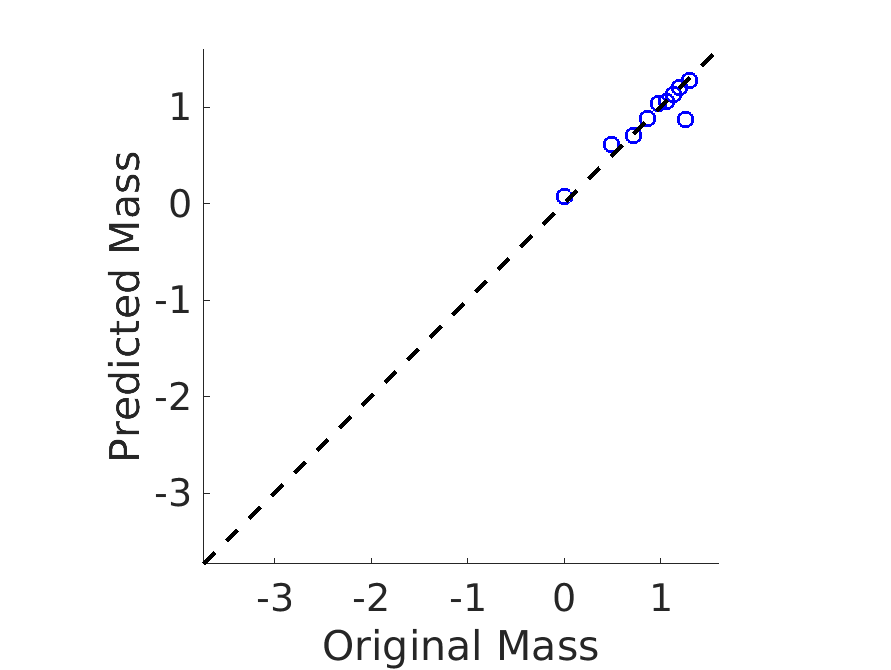}
        \caption{Critical damped system}
        \label{fig:spring_regression_comparison_mass_critical_damped_condition}
    \end{subfigure}
    \hfill
    \begin{subfigure}[t]{0.325\linewidth}
        \centering
        \includegraphics[width=\linewidth]{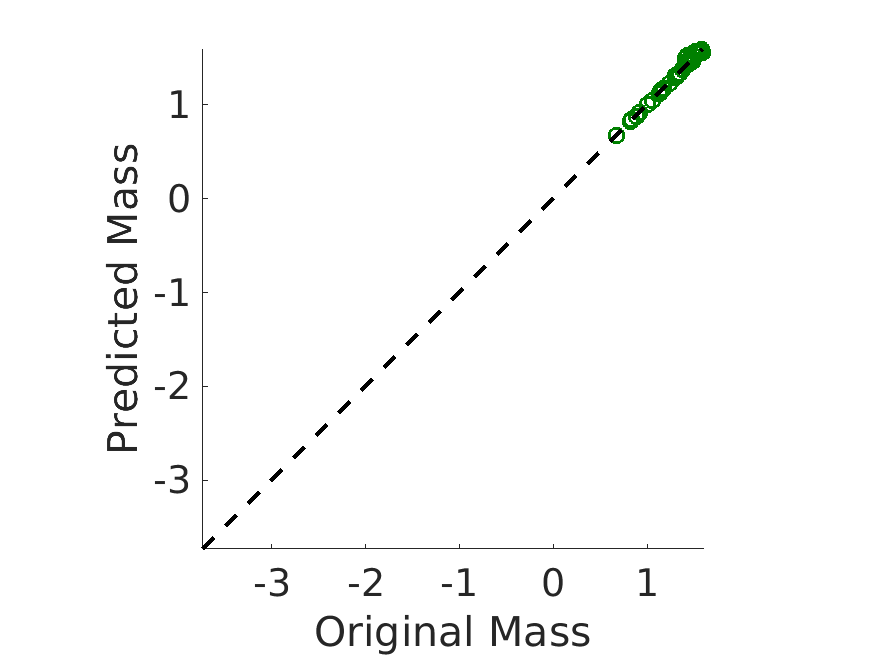}
        \caption{Underdamped system}
        \label{fig:spring_regression_comparison_mass_underdamped_condition}
    \end{subfigure}

    \caption{Regression plots (log scale) comparing original and predicted damping (top row) and mass (bottom row) for overdamped, critically damped, and underdamped spring systems.}
    \label{fig:spring_regression_comparison_all}
\end{figure}

Figure \ref{fig:reg_no_noise} shows the performance of the unsupervised regression algorithm in terms of identification of the system and the surrounding properties represented by the mass and the damping factor, respectively, when there is no noise present in the data. The $R^2$ score observed for the mass and damping factor regression is $1.0$ and $1.0$, respectively. This shows that the model performs efficient discovery of the governing equation if there is no noise present in the system. Figure \ref{fig:spring_regression_comparison_all} shows performance of the model for noisy data. Here, for visualizing the regression results, we use log transformation of both the original and predicted values of the properties. Figure \ref{fig:spring_regression_comparison_damping_factor_overdamped_condition} and \ref{fig:spring_regression_comparison_mass_overdamped_condition} show the results for overdamped conditions for a range of different damping factors and mass values, respectively. We observe that for some cases in the overdamping condition, the model predicts very small mass and damping factor values. Since overdamping condition solution is in the domain of superposition of exponential functions, the algorithm sometime discovers a system which is similar to a first order system. Additionally, these small values values are obtained in such cases where the damping factor or the mass is considered very high. Since we do not restrict the unsupervised discovery problem to a specific order of the underlying governing equation and rather perform a sparse regression problem, due to the exponential nature of the response the model performance degrades in terms of this physical property identification task.





Subsequently Figures \ref{fig:spring_regression_comparison_damping_factor_critical_damped_condition} and \ref{fig:spring_regression_comparison_mass_critical_damped_condition} show the regression results for critical damped system on a logarithmic scale. The $R^{2}$ score for the two regression analysis are $0.6734$ and $0.8849$, respectively. From the metric values we can infer that the model performs relatively better for identifying mass of the system compared to damping factor. This inference becomes more prominent for underdamped response. Figures \ref{fig:spring_regression_comparison_damping_factor_underdamped_condition} and \ref{fig:spring_regression_comparison_mass_underdamped_condition} show the regression performance for underdamped responses. The $R^{2}$ score for underdamped condition for the damping factor and the mass are $0.9758$ and $0.9607$, respectively. This indicates that the model performs significantly more accurate for discovering underlying governing system for underdamped systems compared to overdamped and critical damped responses. Another critical point to notice is that the model performance improves as the impact of mass becomes significant in the system. The performance reduces as the impact of damping becomes significant in the system.

Table \ref{tab:comparative_study_sindy} shows the comparative study between our proposed model and SINDy and its variants. SINDy was implemented using the PySINDy library \cite{de2020pysindy}. The results provided for SINDy was obtained for discovering a single fixed order governing equation for the spring mass system. We observe that the sparse prediction using SINDy to predict sparse coefficient set for a higher order ODE reduces the predictability of SINDy. We also had to fix the order of ODE to be $3$ in SINDy apriori. This is problematic as it is typically unknown. SINDy is more suitable to discovering a system of time dependent first order equation where non homogeneity is present in the system. 
Weak-SINDy \cite{messenger2021weak} is a variant of SINDy focusing on discovery of partial differential equation. From the results we observe that for low noise condition, SINDy with STLSQ is able to perform comparable to our model but as we increase the scale of noise in the dataset its performance degrades. Additionally, Weak-SINDy performs significantly worse for all scales of noise. This is due to the implicit nature of Weak-SINDy which is more suitable for discovering partial differential equations.

\begin{table}[htbp]
\centering
\caption{Model Comparison with State-of-the-art Methods.}
\small
\begin{tabular}{rlrrr}
\toprule
\textbf{Noise} & \textbf{Damping} & \textbf{SINDy-STLSQ} & \textbf{Weak-SINDy} & \textbf{Our Approach} \\
\midrule

\multirow{3}{*}{0.001} 
& Overdamped      & 2.31E-07 & 72.865  & 2.35E-07 \\
& Critical Damped & 1.77E-06 & 142.708 & 1.52E-05 \\
& Underdamped     & 6.71E-05 & 846.31  & 3.05E-06 \\

\midrule

\multirow{3}{*}{0.01} 
& Overdamped      & 1.63E-06 & 43.683  & 1.60E-07 \\
& Critical Damped & 6.43E-05 & 9.396   & 3.05E-06 \\
& Underdamped     & 2.95E-04 & 26.606  & 1.55E-05 \\

\midrule

\multirow{3}{*}{0.1} 
& Overdamped      & 1.07E-04 & 0.328   & 3.05E-06 \\
& Critical Damped & 1.49E-06 & 27.700  & 0.000152 \\
& Underdamped     & 3.07E-06 & 38.544  & 2.35E-07 \\

\bottomrule
\end{tabular}
\label{tab:comparative_study_sindy}
\end{table}

\begin{figure}[htbp]
    \centering
    \includegraphics[width=0.75\linewidth]{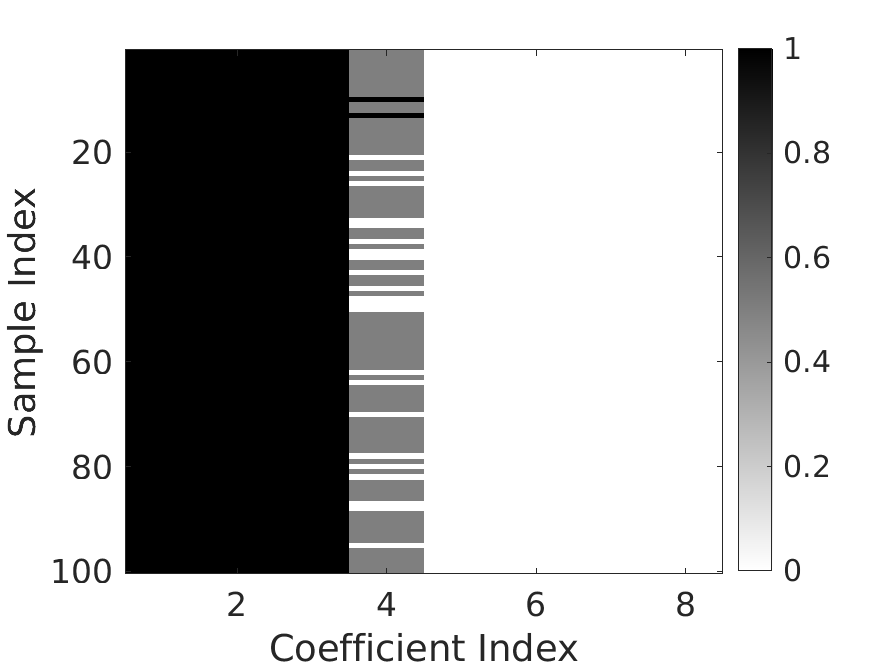}
    \caption{Sparsity in model prediction.}
    \label{fig:sparsity_linear_model}
\end{figure}

Figure \ref{fig:sparsity_linear_model} shows the sparsity in the prediction. All values less than $10^{-4}$ are considered to be $0$ and all values more than $0.98$ are considered to be $1$. Following this operation, all values are normalized between $0$ and $1$. From Figure \ref{fig:sparsity_linear_model}, we observe that most of the prediction is restricted to second-order dynamics and the higher-order terms are not considered by our prediction model. One thing to note, all the analyses are performed without use of any regularization techniques. So, the sparsity observed in the model prediction is inherent property of the model due to linearly independent basis values generated through the polynomial approximation. 

\begin{table}[]
\centering
\caption{Correlation between errors and hyperparameters.}
\begin{tabular}{@{}lll@{}}
\toprule
Parameter           & Approximate General Solution Error & ODE Generation Error \\ \midrule
Population Size     & -0.06485             & 0.02918    \\
Maximum Generations & -0.02482            & 0.16310    \\
K                   & 0.00926              & 0.00354    \\
Noise               & 0.02950              & -0.01605   \\
T                   & 0.15222              & -0.02035   \\ \bottomrule
\end{tabular}
\label{tab:error_hyperparameter_correlation}
\end{table}

Table \ref{tab:error_hyperparameter_correlation} shows the correlation between the approximate general solution error and the ODE generation error of the model. We varied the hyperparameters of the genetic algorithm, such as population size, maximum generations, maximum terms considered in functional form, noise present in the data, and the time period of generation for the dataset within the bounds of $(200, 750), (300, 700), (4, 8), (10^{-5}, 10^{-1}) (10, 50)$, respectively. Most of the hyperparameters have low influence on the model performance. All of the values ranges from $10^{-3}$ to $0.15$. 
\noindent The maximum order of generation has very low influence on the model performance indicating that the model is robust even if more terms are considered. This increases the flexibility of the model for sparse prediction of the governing system considering the structure of a high order general ODE (order $\geq 5$). Also, influence of noise is low on the model performance which indicates that the model is robust to noise present in the dataset. This shows the capability of the model to provide interpretable discovery under high noise condition.

\subsection{Case Study 2: Photolytic degradation of Estrogenic Chemicals}

\begin{figure}[htbp]
    \centering

    \begin{subfigure}[t]{0.49\linewidth}
        \centering
        \includegraphics[width=\linewidth]{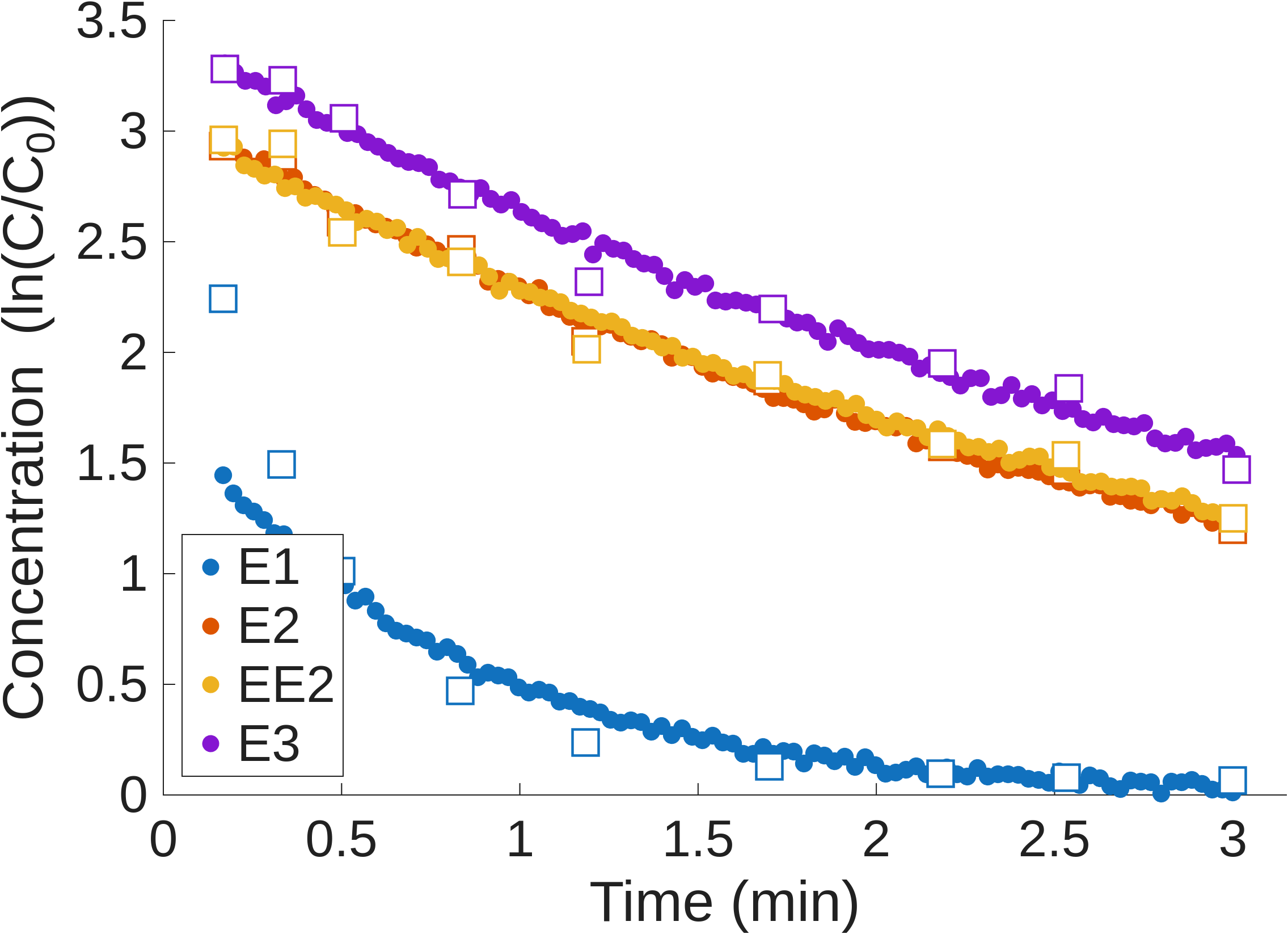}
        \caption{UVA Photolysis Profile}
        \label{fig:photolysis_UVA_profile}
    \end{subfigure}
    \hfill
    \begin{subfigure}[t]{0.49\linewidth}
        \centering
        \includegraphics[width=\linewidth]{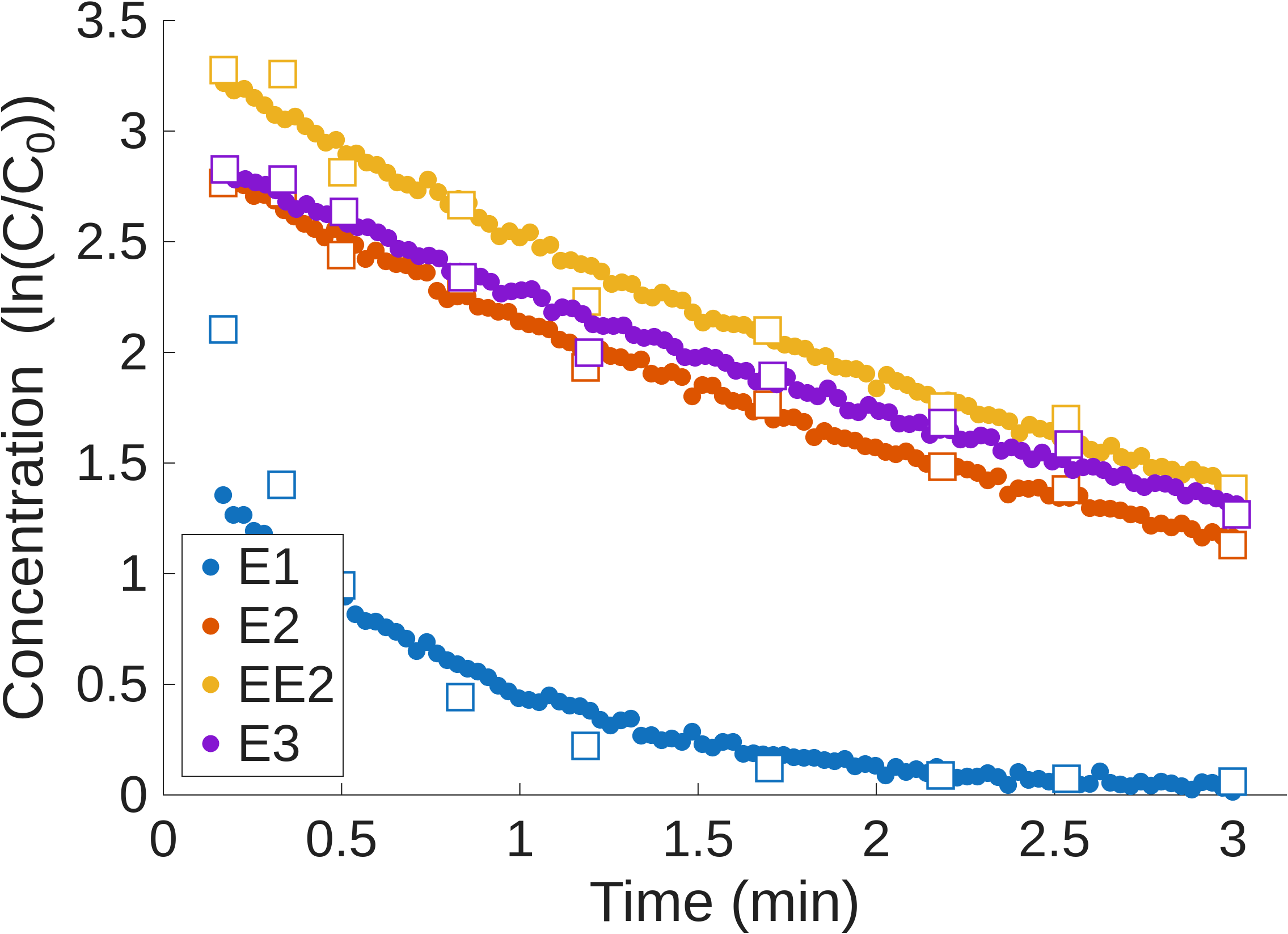}
        \caption{UVC Photolysis Profile}
        \label{fig:photolysis_UVC_profile}
    \end{subfigure}
    \caption{Concentration profiles for different EDC components under photolysis.}
    \label{fig:edc_concentration_profile}
\end{figure}

\begin{table}[htbp]
\centering
\caption{Comparison of original and predicted ODEs for each component.}
\label{tab:ode_comparison_detailed}
\begin{tabular}{lll}
\hline
Component & 1st-order ODE & Predicted ODE \\ \hline
UVA-E1    & $\frac{dC}{dt} + 0.2240\,C = 0$ 
          & $\frac{dC}{dt} + 0.2345\,C - 0.0025\,\frac{d^2C}{dt^2} = 0$ \\
UVA-E2    & $\frac{dC}{dt} + 0.0744\,C = 0$ 
          & $\frac{dC}{dt} + 0.0843\,C + 0.0046\,\frac{d^2C}{dt^2} = 0$ \\
UVA-EE2   & $\frac{dC}{dt} + 0.0171\,C = 0$ 
          & $\frac{dC}{dt} + 0.0142\,C + 0.0025\,\frac{d^2C}{dt^2} = 0$ \\
UVA-E3    & $\frac{dC}{dt} + 0.0959\,C = 0$ 
          & $\frac{dC}{dt} - 0.0859\,C - 0.0019\,\frac{d^2C}{dt^2} = 0$ \\
UVC-E1    & $\frac{dC}{dt} + 1.3000\,C = 0$ 
          & $\frac{dC}{dt} + 1.2700\,C + 0.0007\,\frac{d^2C}{dt^2} = 0$ \\
UVC-E2    & $\frac{dC}{dt} + 0.3054\,C = 0$ 
          & $\frac{dC}{dt} + 0.3090\,C + 0.0020\,\frac{d^2C}{dt^2} = 0$ \\
UVC-EE2   & $\frac{dC}{dt} + 0.3054\,C = 0$ 
          & $\frac{dC}{dt} + 0.2690\,C + 0.0014\,\frac{d^2C}{dt^2} = 0$ \\
UVC-E3    & $\frac{dC}{dt} + 0.3054\,C = 0$ 
          & $\frac{dC}{dt} + 0.2940\,C + 0.0020\,\frac{d^2C}{dt^2} = 0$ \\
\hline
\end{tabular}
\end{table}

\begin{table}[]
\centering
\caption{Rate constant prediction comparison.}
\begin{tabular}{lrrr}
\hline
Component & \makecell{1st-order\\Rate Constant} & \makecell{Predicted\\Rate Constant} & \makecell{MSE} \\ \hline
UVA-E1    & 0.22400                             & 0.23900                              & 0.00023 \\
UVA-E2    & 0.07440                             & 0.08430                              & 0.00010 \\
UVA-EE2   & 0.01710                             & 0.01420                              & 0.00001 \\
UVA-E3    & 0.09590                             & 0.08590                              & 0.00010 \\
UVC-E1    & 1.30000                             & 1.27000                              & 0.00090 \\
UVC-E2    & 0.30543                             & 0.30900                              & 0.00001 \\
UVC-EE2   & 0.30543                             & 0.26900                              & 0.00133 \\
UVC-E3    & 0.30543                             & 0.29400                              & 0.00013 \\ \hline
\end{tabular}
\label{tab:kinetic_model_comparison}
\end{table}

\begin{figure}[htbp]

    \centering

    \begin{subfigure}[t]{0.49\linewidth}
        \centering
        \includegraphics[width=\linewidth]{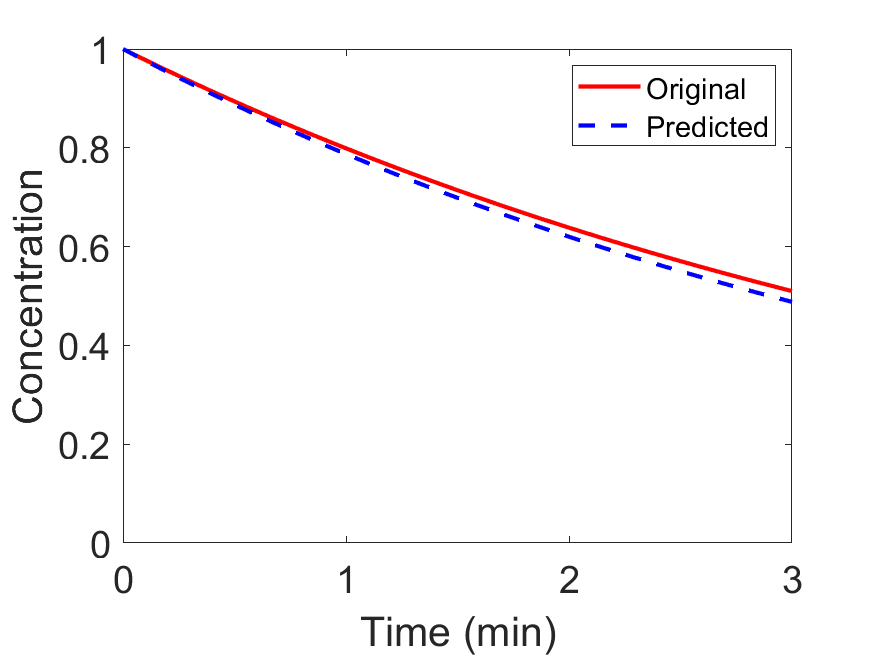}
        \caption{E1 (Estrone) Degradation}
        \label{fig:uva_e1_degradation}
    \end{subfigure}
    \hfill
    \begin{subfigure}[t]{0.49\linewidth}
        \centering
        \includegraphics[width=\linewidth]{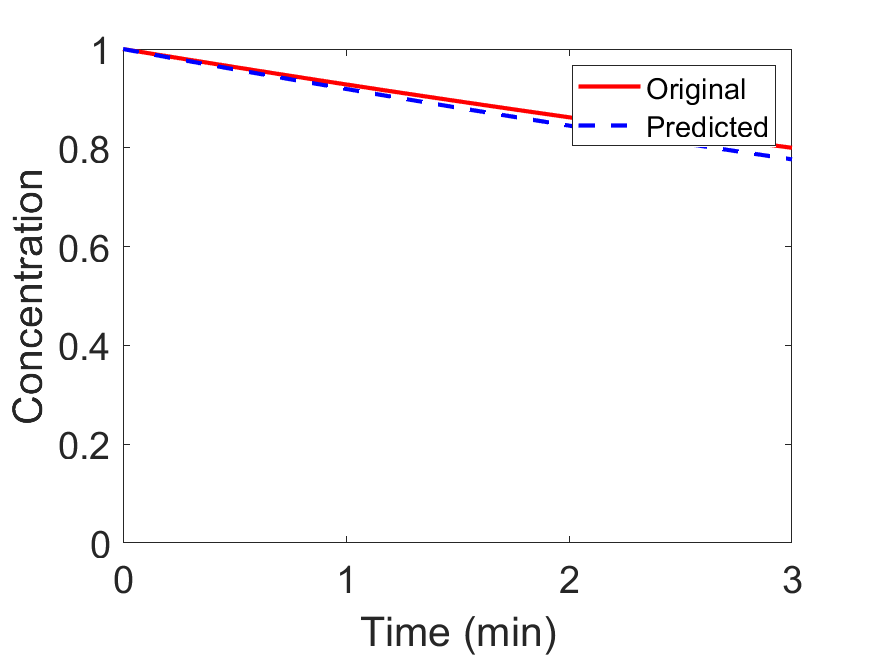}
        \caption{E2 ($17\text{-}\beta$ Estradiol Degradation}
        \label{fig:uva_e2_degradation}
    \end{subfigure}
    \hfill
    \begin{subfigure}[t]{0.49\linewidth}
        \centering
        \includegraphics[width=\linewidth]{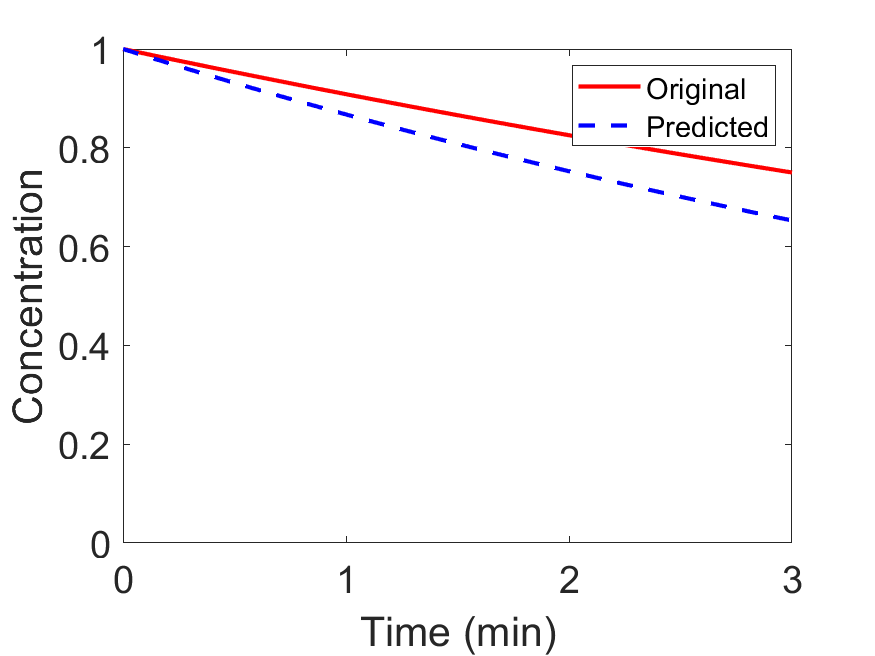}
        \caption{EE2 ($17\text{-}\alpha$ Ethynylestradiol) Degradation}
        \label{fig:uva_ee2_degradation}
    \end{subfigure}
    \hfill
    \begin{subfigure}[t]{0.49\linewidth}
        \centering
        \includegraphics[width=\linewidth]{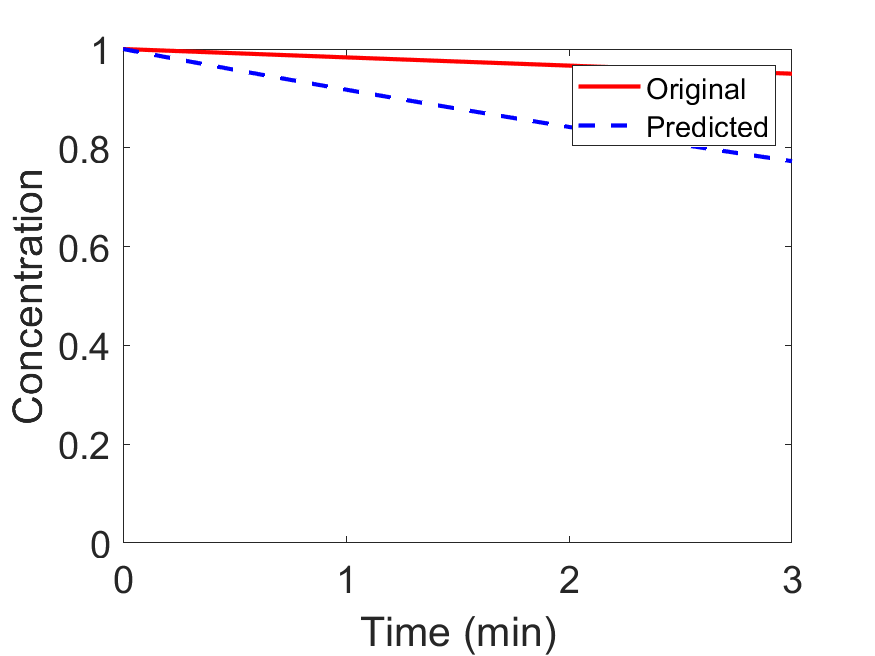}
        \caption{E3 (Estriol) Degradation}
        \label{fig:uva_e3_degradation}
    \end{subfigure}

    \caption{Comparison of original and predicted degradation profiles for EDC components under UVA photolysis.}
    \label{fig:degradation_comparison_uva_photolysis}
\end{figure}

\begin{figure}[htbp]
    \centering

    \begin{subfigure}[t]{0.49\linewidth}
        \centering
        \includegraphics[width=\linewidth]{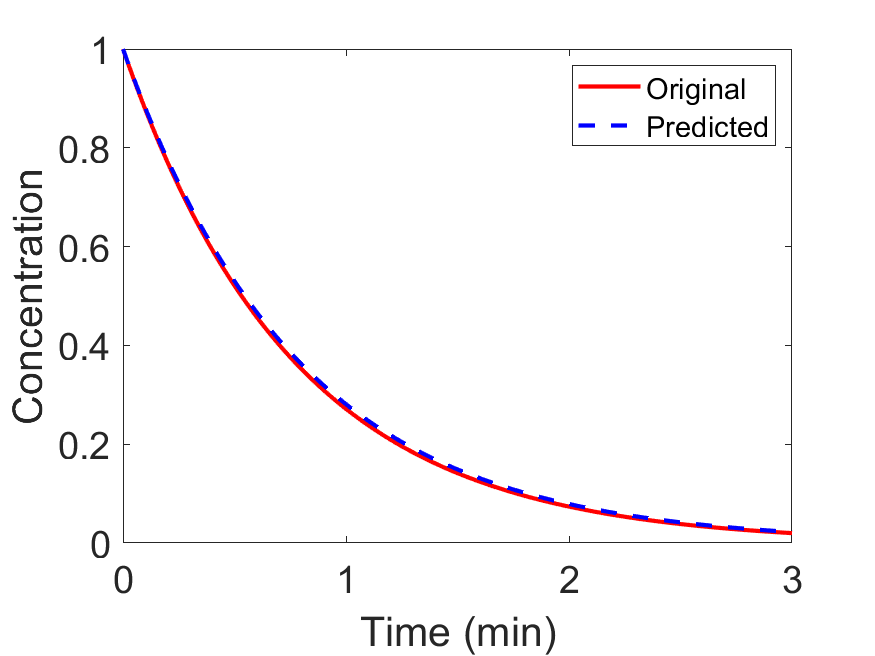}
        \caption{E1 (Estrone) Degradation}
        \label{fig:uvc_e1_degradation}
    \end{subfigure}
    \hfill
    \begin{subfigure}[t]{0.49\linewidth}
        \centering
        \includegraphics[width=\linewidth]{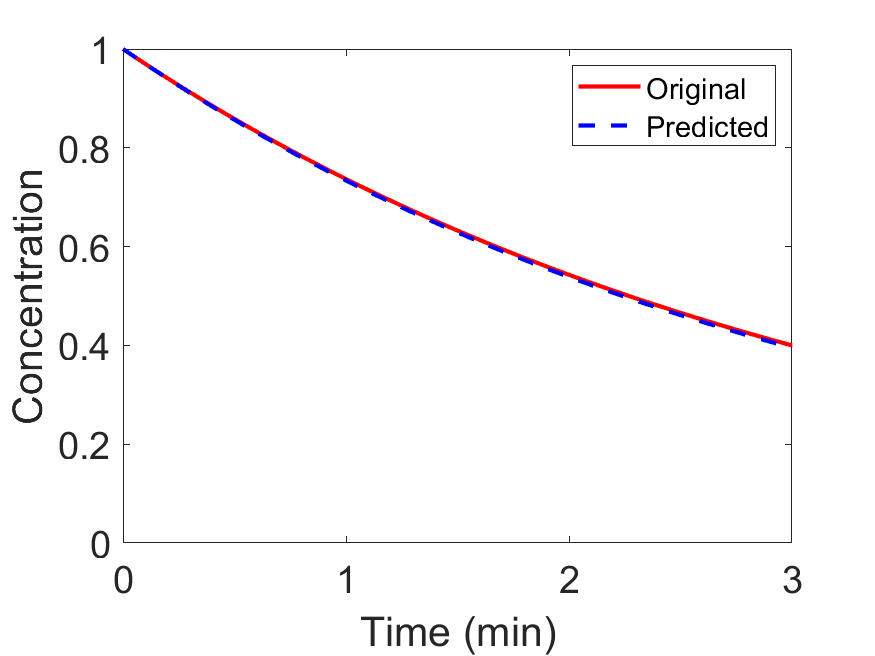}
        \caption{E2 ($17\text{-}\beta$ Estradiol) Degradation}
        \label{fig:uvc_e2_degradation}
    \end{subfigure}
    \hfill
    \begin{subfigure}[t]{0.49\linewidth}
        \centering
        \includegraphics[width=\linewidth]{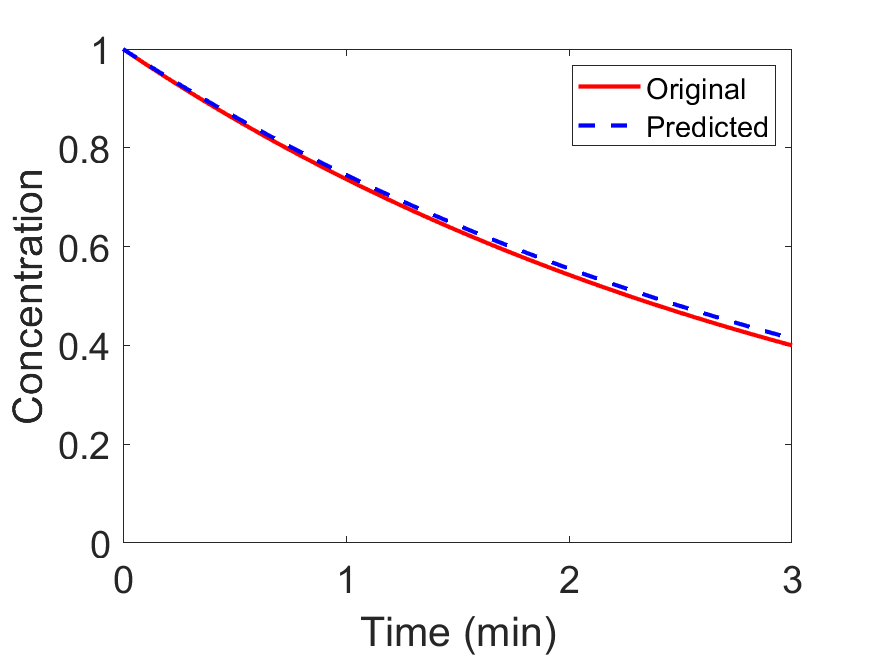}
        \caption{EE2 ($17\text{-}\alpha$ Ethynylestradiol) Degradation}
        \label{fig:uvc_ee2_degradation}
    \end{subfigure}
    \hfill
    \begin{subfigure}[t]{0.49\linewidth}
        \centering
        \includegraphics[width=\linewidth]{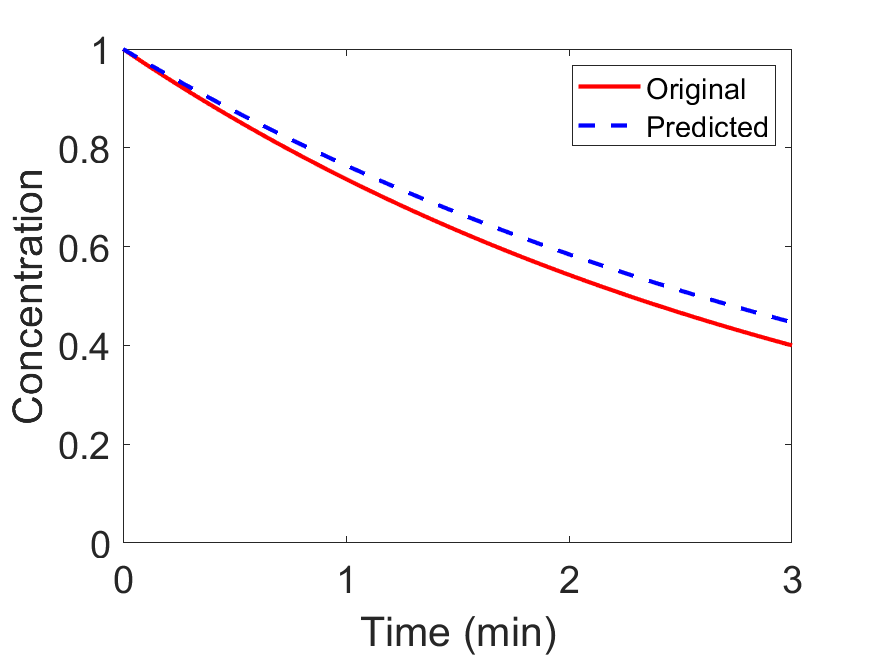}
        \caption{E3 (Estriol) Degradation}
        \label{fig:uvc_e3_degradation}
    \end{subfigure}

    \caption{Comparison of original and predicted degradation profiles for EDC components under UVC photolysis.}
    \label{fig:degradation_comparison_uvc_photolysis}
\end{figure}

\begin{figure}[htbp]
    \centering
    \includegraphics[width=0.75\linewidth]{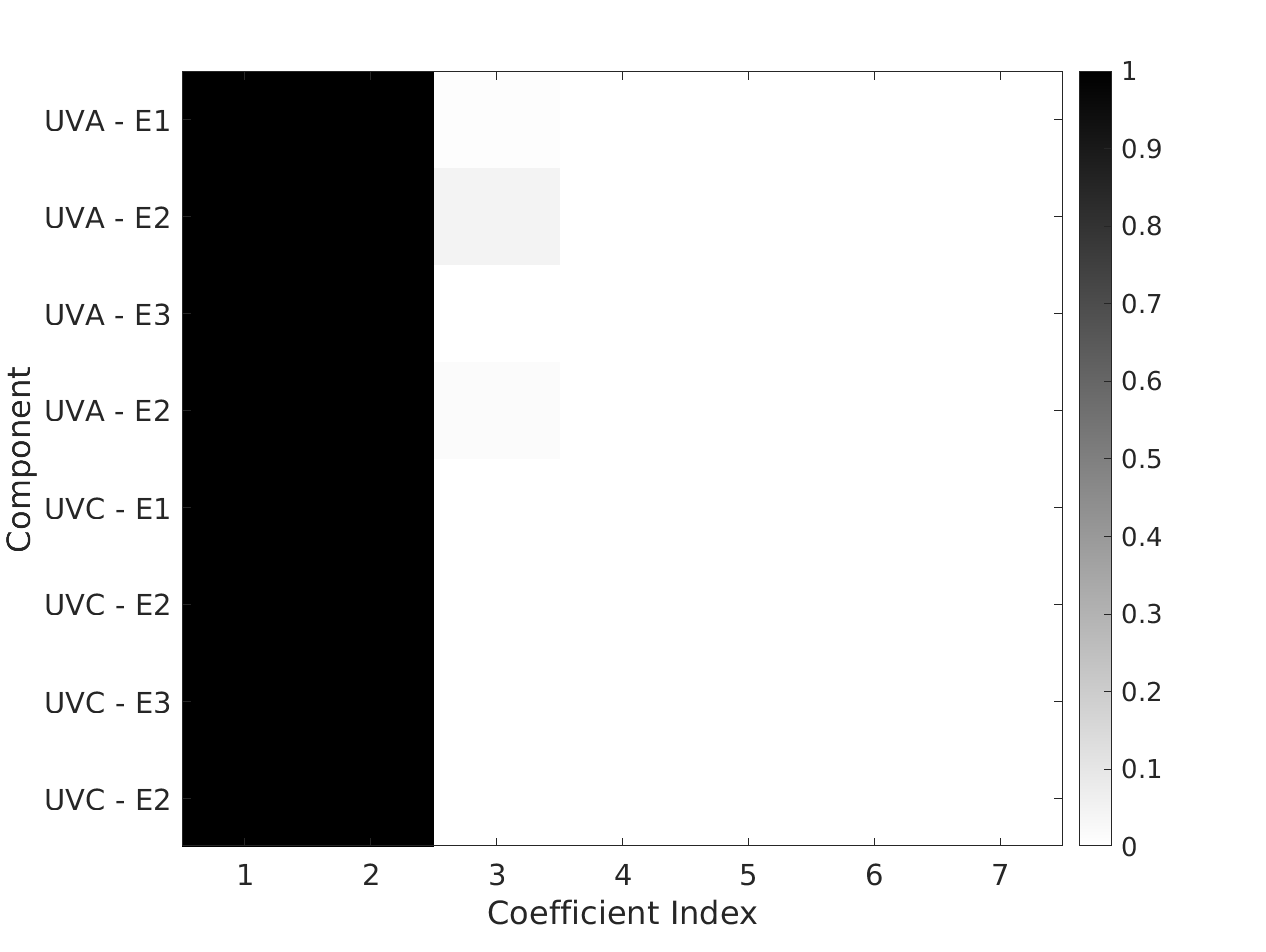}
    \caption{Sparsity in model prediction.}
    \label{fig:sparsity_kinetic_model}
\end{figure}

In this case study, we consider model identification from an experimental study of the photolytic degradation of mixtures of EDCs (Estrogen Disrupting chemicals). EDCs are hazardous chemicals, and their release via degradation, even in low concentrations, can induce adverse health and environmental effects including sexual reversal and population imbalance on aquatic species. To that end, Puma et al.\cite{puma2010photocatalytic} performed a detailed kinetic study of the photolytic and photocatalytic degradation of mixtures of EDCs. Their analysis suggested that the degradation process could be approximated by apparent first-order kinetics. Our objectives are twofold: (i) to derive ODEs that describe the time evolution of EDC concentrations using real experimental data, and (ii) to assess and potentially refine the first-order kinetic assumption through data-driven modeling. This also serves to evaluate the interpretability and reliability of the discovered models through our proposed approach.

Figures \ref{fig:photolysis_UVA_profile} and \ref{fig:photolysis_UVC_profile} show the logarithmic concentration profiles of the various EDC chemicals (E1, E2, EE2, and E3) during UVA and UVC light based photolysis, respectively. From the profiles of UVA and UVC photolysis, it can be presumed that the conversion of the all the chemicals under photolysis is low. Furthermore, all the chemicals, except E1, might have followed a first order kinetics. For E1, the degradation profile deviates from first order kinetics over longer irradiation and achieves a high conversion. This is counter intuitive when compared with conversion of other chemicals. 


For these types of kinetic studies, we first find the $\alpha, \beta, \gamma$ parameters of the approximate general solution of the data. The initial concentration of each chemical in the mixture is given to us. The number of data points in the experimental observation is too small to include it into a genetic algorithm framework. For this reason, we first considered a logarithmic transformation of the normalized concentrations with respect to the initial concentration. Following this transformation, we added $1000$ new data samples based on a linear approximation of the experimental data points. Subsequently, the inverse transformation is performed on the data and a random noise of scale $0.001$ is added to the data to obtain the training dataset. We use the entire dataset since the purpose is to discover a mathematical description of the underlying system.

Figure \ref{fig:degradation_comparison_uva_photolysis} shows the data used for model identification for the UVA photolysis of the EDC components in the mixture, where the square-marked points represent the samples obtained from actual experiments, and the solid circles represent the added data points through linear approximation. From these data, we observe that $E1$ and $E2$ have similar response. But for $EE2$ and $E3$ the model discovers a system which deviates from the original system. The conversions of both $EE2$ and $E3$ are found to be low compared to conversions of $E1$ and $E2$. For low conversion system, where the rate constant is of order $10^{-2}$ the model performance reduces. 

Table \ref{tab:ode_comparison_detailed} shows the comparison of the assumed first-order ODEs, as assumed by Puma et al., and the predicted ODEs obtained from our data-driven approach. Higher order terms less than $10^{-5}$ are neglected in the predicted ODEs. This can also be observed from Table \ref{tab:kinetic_model_comparison} which shows the comparison between original rate constants derived from the apparent first order kinetics and the predicted rate constants. The predicted rate constants are close to the original rate constants for all the components of of EDC under UVA irradiation.

Figure \ref{fig:degradation_comparison_uvc_photolysis} shows the results for UVC photolysis. Compared to UVA irradiation, where the conversion of all the components are relatively low, our prediction is better for UVC systems with higher conversion. From Table  \ref{tab:kinetic_model_comparison}, we observe that the model is able to discover the underlying system by predicting the rate constants of the system efficiently. Additionally, the mean square errors for each components are low and in the order of $10^{-4}$.For high conversion system, our approach is able to predict the concentration profile efficiently along with discovering the kinetic rate constant of the system.

Figure \ref{fig:sparsity_kinetic_model} shows the sparsity in the prediction for both irradiations of the EDC mixture. During model discovery, our motivation was not to restrict the model to estimate coefficients of a first order ODE system. Rather we have predicted coefficients for a $P^{th}$ (here $P=7$) order system and observe if the model is still able to discover the order of the underlying system. It is known to us that the system follows an apparent first order kinetic expression. For visualizing the sparsity of the prediction we have considered a specific transformation. We considered all the coefficients predicted under a specific threshold value $\epsilon$ (here $\epsilon = 10^{-4}$) to be $0$ and all the values predicted over $0.1$ to be $1$. Subsequently, we normalize the entire set of predicted values between $0$ to $1$. From the figure, we observe that the model predicts a sparse system for all the cases of irradiation under both UVA and UVC irradiation. This shows the model's capability to perform sparse prediction for a long range of system order consideration. We have not included any regularization technique for this study. This highlights that, if the evaluated basis of the gradient matrix are linearly independent of each other, then the discovered model selects only those gradient terms that is present in the response of the underlying system.

\section{Conclusions}
\label{sec:conclusion}

We presented a data-driven methodology that combines smooth, analytical functional form with spline approximation to discover the governing ordinary differential equations (ODEs) of physical systems. Unlike many existing approaches that rely on fixed basis functions and often produce models lacking physical meaning, our method builds interpretable models that better reflect the underlying dynamics. The smooth approximation enables the accurate computation of gradients, which are linearly independent and serve as the basis for estimating ODE coefficients through a linear system. One of the strengths of our approach is that it avoids the need for explicit regularization while naturally promoting sparsity by emphasizing lower-order terms. We tested the method on two systems: a spring-mass system and a photocatalytic EDC decomposition process. For the spring-mass system, the model captured the dynamics effectively, particularly in underdamped regimes where first-order terms dominate. In the EDC case, the method successfully identified rate constants, performing better in high-conversion scenarios. While standard linear models perform well with noise-free data, they degrade significantly in the presence of even small noise. In contrast, our approach remains robust, making it suitable for real-world experimental data. However, we also observed limitations when applying the method to higher-order or variable-coefficient systems, which are less common but still relevant in some engineering applications. In future work, we plan to develop a  framework that integrates this methodology into a user-friendly pipeline, enabling broader application across chemical and other engineering domains.

\section*{Acknowledgement} 
The authors gratefully acknowledge partial funding support from the U.S. National Science Foundation (NSF CAREER award CBET-1943479) and the U.S. Environmental Protection Agency (EPA) Project Grant 84097201.

\appendix
\section{Appendix A: Spring Mass Data Generation}
\label{appx:spring_mass_data_generation}
The spring system is represented by the following differential equation.



\begin{ceqn}
\label{eq:spring_eq}
    \begin{equation}
    m\frac{d^2x(t)}{dt^2} + b\frac{dx(t)}{dt} + kx(t) = 0
\end{equation}
\end{ceqn}

\noindent where, $m\frac{d^2x(t)}{dt^2}$ represents the inertial term that resists acceleration. $m$ is the mass of the block. $b\frac{dx(t)}{dt}$ corresponds to the damping force that resists motion and is proportional to the velocity. Here, $k$ represents the spring constant. $kx(t)$ shows the restoring force exerted by the spring. The boundary conditions of the ODE in Eq. \ref{eq:spring_eq} are given by

\begin{ceqn}
    \begin{equation}
    x(t=0) = x_{0}
\end{equation}
\end{ceqn}

\begin{ceqn}
    \begin{equation}
    \frac{dx(t=0)}{dt} = x_{1}.
\end{equation}
\end{ceqn}

\noindent $d$ can be calculated from the following equation $d = b^2 - 4mk$ which represents the parameter used to solve the differential equation and gives rise to different scenarios given by

\begin{ceqn}
\begin{equation}
\text{(i) } d > 0 \quad \text{(ii) } d = 0 \quad \text{(iii) } d < 0
\end{equation}
\end{ceqn}

\textbf{Overdamped Response:}
Overdamped response signifies that the system does not perform any oscillations due to the high frictional force acting on the body forcing it to reach steady state before oscillation. Upon solving for general solution, we get the following form:

\begin{ceqn}
    \begin{equation}
    x(t) = e^{\frac{-b}{2m}t}(C_{1}e^{\frac{\sqrt{d}}{2m}t} + C_{2}e^{\frac{-\sqrt{d}}{2m}t})
\end{equation}
\end{ceqn}

Solving it for the following boundary conditions:

\begin{ceqn}
\begin{align*}
    x(t=0) &= x_{0}, \quad \\
    \frac{dx(t=0)}{dt} &= x_{1}
\end{align*}
\end{ceqn}

We get

\begin{ceqn}
\begin{align*}
    C_{1} &= \left(x_{1} + \frac{b}{2m}x_{0}\right) \cdot \frac{m}{\sqrt{d}}, \\
    C_{2} &= -x_{1} \cdot \frac{m}{\sqrt{d}} + x_{0} \cdot \left(1 - \frac{b}{2\sqrt{d}}\right).
\end{align*}
\end{ceqn}

\textbf{Critical Damped Response:}
 Physically it signifies the threshold between the non-oscillatory and oscillatory response of the system. Introducing more damping to such system would result in a non-oscillatory response (case 1) and reducing damping in these systems would introduce oscillations into the system (case 3). Upon solving the differential equation, we obtain the following response:

\begin{ceqn}
    \begin{equation}
    x(t) = e^{\frac{-b}{2m}t}(C_{1} + C_{2}t)
    \end{equation}
\end{ceqn}

Solving it for the previously mentioned boundary conditions, we get

\begin{ceqn}
    \begin{align*}
    C_{0} &= x_{0}, \quad \\
    C_{1} &= x_{1} + \frac{b}{2m}x_{0}.
\end{align*}
\end{ceqn}

\textbf{Underdamped Response:}
In a physical sense, it indicates that the system is governed by oscillatory response. Upon decreasing the damping of the system the frequency of the oscillations increases. Upon solving differential equation we get the following solution:

\begin{ceqn}
    \begin{equation}
    x(t) = e^{\frac{-b}{2m}t}(C_{1}cos(\frac{\sqrt{|d|}}{2m}t) + C_{2}sin(\frac{\sqrt{|d|}}{2m}t))
\end{equation}
\end{ceqn}

Solving it for the previously mentioned boundary conditions we get,

\begin{ceqn}
    \begin{align*}
    C_{1} &= x_{0}(\frac{1}{2} + \frac{b}{2\sqrt{|d|}}) + \frac{mx_{1}}{\sqrt{|d|}}, \quad \\
    C_{2} &= x_{0}(\frac{1}{2} - \frac{b}{2\sqrt{|d|}}) - \frac{mx_{1}}{\sqrt{|d|}}
\end{align*}
\end{ceqn}

\bibliographystyle{unsrt}  
\bibliography{references}

\end{document}